\documentclass[conference]{IEEEtran}
\IEEEoverridecommandlockouts
\usepackage{cite}
\usepackage{amsmath,amssymb,amsfonts}
\usepackage{algorithmic}
\usepackage{graphicx}
\usepackage{textcomp}
\usepackage{xcolor}
\def\BibTeX{{\rm B\kern-.05em{\sc i\kern-.025em b}\kern-.08em
    T\kern-.1667em\lower.7ex\hbox{E}\kern-.125emX}}
\usepackage{booktabs}
\usepackage{svg}
\usepackage{xspace}
\usepackage[acronym]{glossaries}
\makeglossaries
\usepackage{url}
\usepackage{color}
\usepackage{multirow}
\usepackage{enumitem}
\usepackage{array}
\usepackage[nolist,nohyperlinks]{acronym}
\usepackage{footnote}
\makesavenoteenv{tabular}
\usepackage{hyperref}
\usepackage{multirow}
\usepackage{tablefootnote}
 
\newcommand{\resunetplusplus}{ResUNet++\xspace}
\acrodef{GI}{gastrointestinal}
\acrodef{CNN}{Convolutional Neural Network}
\acrodef{FCM}{Fuzzy C-mean Clustering}
\acrodef{ROI}{Region of Interest}
\acrodef{AI}{Artificial Intelligence} 
\acrodef{CAI}{Computer Assisted Intervention} 
\acrodef{ML}{machine learning}
\acrodef{DL}{Deep Learning}
\acrodef{IOU}[IoU]{Intersection over Union}
\acrodef{ROI}{Region of Interest}
\acrodef{CRC}{Colorectal Cancer}
\acrodef{CADx}{Computer-Aided Diagnosis}
\acrodef{WCE}{Wireless Capsule Endoscopy}  
\acrodef{BoF}{Bag of Feature}  
\acrodef{GIANA}{Gastrointestinal Image ANAlysis} 
\acrodef{FCNN}{Fully Convolutional Neural Network}
\acrodef{FCN}{Fully Convolutional Network}
\acrodef{ASPP}{Atrous Spatial Pyramidal Pooling }
\acrodef{SGDR}{Stochastic Gradient Descent with Restart}
\acrodef{AUC-ROC}{Area Under Curve - Receiver Operating Characteristic}
\acrodef{ROC}{receiver operating characteristic}
\acrodef{MSE}{Mean Square Error} 
\acrodef{SGD}{Stochastic Gradient Descent}
\acrodef{BoF}{bag of feature}
\acrodef{NLP}{Natural Language Processing}
\acrodef{MSE}{Mean Square Error}
\acrodef{mIoU}{mean Intersection over Union}
\acrodef{ReLU}{Rectified Linear Unit}
\acrodef{CRC}{colorectal cancer}
\acrodef{BN}{Batch Normalization}
\acrodef{SE}{squeeze and excitation}
\acrodef{TTA}{Test-Time Augmentation }
\acrodef{CRF}{Conditional Random Field}
\acrodef{ASPP}{atrous spatial pyramid pooling}
\acrodef{mIoU}{mean Intersection over Union}
\acrodef{AUC}{area under curve}
\acrodef{SD}{standard definition}
\acrodef{GIANA}{Gastrointestinal Image ANAlysis}
\acrodef{DSC}{Dice coefficient}
\acrodef{DSC}{Dice coefficient}
\acrodef{SOTA}{State-of-the-art}
\begin{document}

\title{A Comprehensive Study on Colorectal Polyp Segmentation with \resunetplusplus, Conditional Random Field and Test-Time Augmentation}

\author{Debesh Jha, Pia H. Smedsrud, Dag Johansen, Thomas de Lange, 
H{\aa}vard D. Johansen, \\P{\aa}l Halvorsen, and Michael A. Riegler

\protect\thanks{A preliminary version of this paper was presented in~\cite{jha2019resunet++}.}
\thanks{Manuscript received xxxx-xx-xx; revised xxxx-xx-xx; accepted xxxx-xx-xx; Date of Publication xxxx-xx-xx. } 
\thanks{The authors are with the SimulaMet, Norway,  Augere Medical AS, Norway,  UiT The Arctic University of Norway, University of Oslo, Norway, Oslo Metropolitan University, Norway, Sahlgrenska University Hospital, Mölndal, Sweden, and Bærum Hospital, Vestre Viken, Norway (Corresponding author: Debesh Jha (e-mail: debesh@simula.no))}}

\maketitle
\begin{abstract}
Colonoscopy is considered the gold standard for detection of colorectal cancer and its precursors. Existing examination methods are, however, hampered by high overall miss-rate, and many abnormalities are left undetected.
Computer-Aided  Diagnosis systems based on advanced machine learning algorithms are touted as a game-changer that can identify regions in the colon overlooked by the physicians during endoscopic examinations, and help detect and characterize lesions. 
In previous work, we have proposed the \resunetplusplus architecture and demonstrated that it produces more efficient results compared with its counterparts U-Net and ResUNet. In this paper, we demonstrate that further improvements to the overall prediction performance of the \resunetplusplus architecture can be achieved by using \ac{CRF} and \ac{TTA}. We have performed extensive evaluations and validated the improvements using six publicly available datasets: Kvasir-SEG, CVC-ClinicDB, CVC-ColonDB, ETIS-Larib Polyp DB, ASU-Mayo Clinic Colonoscopy Video Database, and CVC-VideoClinicDB. Moreover, we compare our proposed architecture and resulting model with other State-of-the-art  methods. To explore the generalization capability of \resunetplusplus on different publicly available polyp datasets, so that it could be used in a real-world setting, we performed an extensive cross-dataset evaluation. The experimental results show that applying \ac{CRF} and \ac{TTA} improves the performance on various polyp segmentation datasets both on the same dataset and cross-dataset. To check the model's performance on difficult to detect polyps, we selected, with the help of an expert gastroenterologist, $196$ sessile or flat polyps that are less than ten millimeters in size. This additional data has been made  available as a subset of Kvasir-SEG. Our approaches showed good results for flat or sessile and smaller polyps, which are known to be one of the major reasons for high polyp miss-rates. This is one of the significant strengths of our work and indicates that our methods should be investigated further for use in clinical practice.

\end{abstract}

\begin{IEEEkeywords}
Colonoscopy, polyp segmentation, \resunetplusplus, conditional random field, test-time augmentation, generalization 
\end{IEEEkeywords}
\vspace{-4mm}
\section{Introduction}
Cancer is a primary health problem of contemporary society, with \ac{CRC} being the third most prevailing type in terms of cancer incidence and second in terms of mortality globally~\cite{bray2018global}. Colorectal polyps are the precursors for the \ac{CRC}. Early detection of polyps through high-quality colonoscopy and regular screening are cornerstones for the prevention of colorectal cancer~\cite{matsuda2017advances}, since adenomas can be found and resected before transforming to cancer and subsequently reducing \ac{CRC} morbidity and mortality. 

\begin{figure} [t]
    \centering
    \includegraphics[width=1.7cm, height=1.7cm]{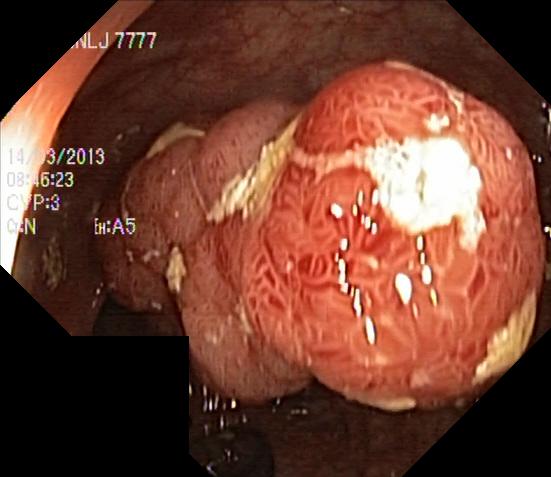}
    \includegraphics[width=1.7cm, height=1.7cm]{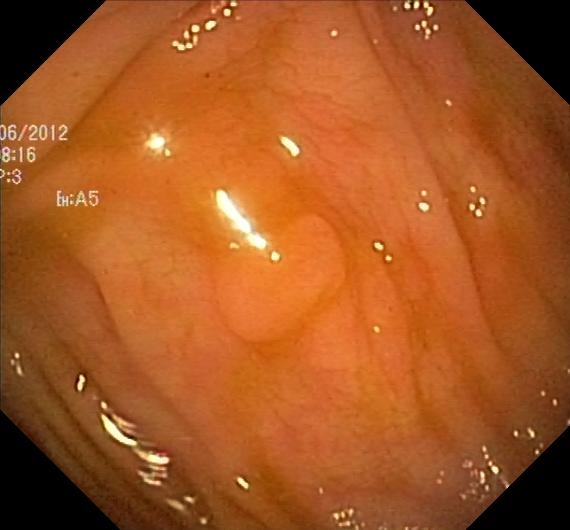}
    \includegraphics[width=1.7cm, height=1.7cm]{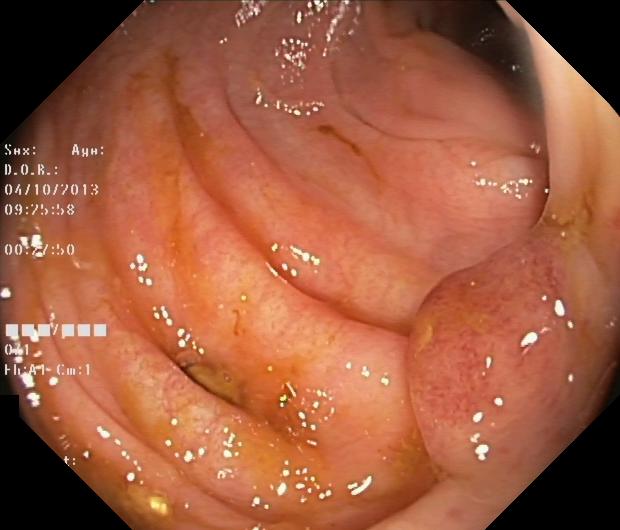}
    \includegraphics[width=1.7cm, height=1.7cm]{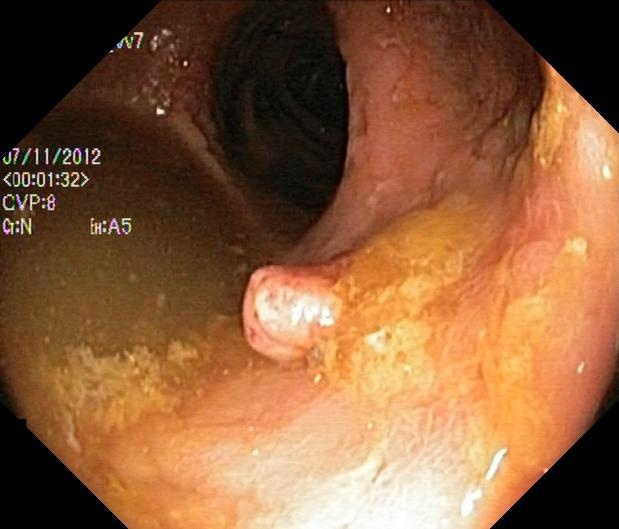}\\
    \vspace{1mm}
    \includegraphics[width=1.7cm, height=1.7cm]{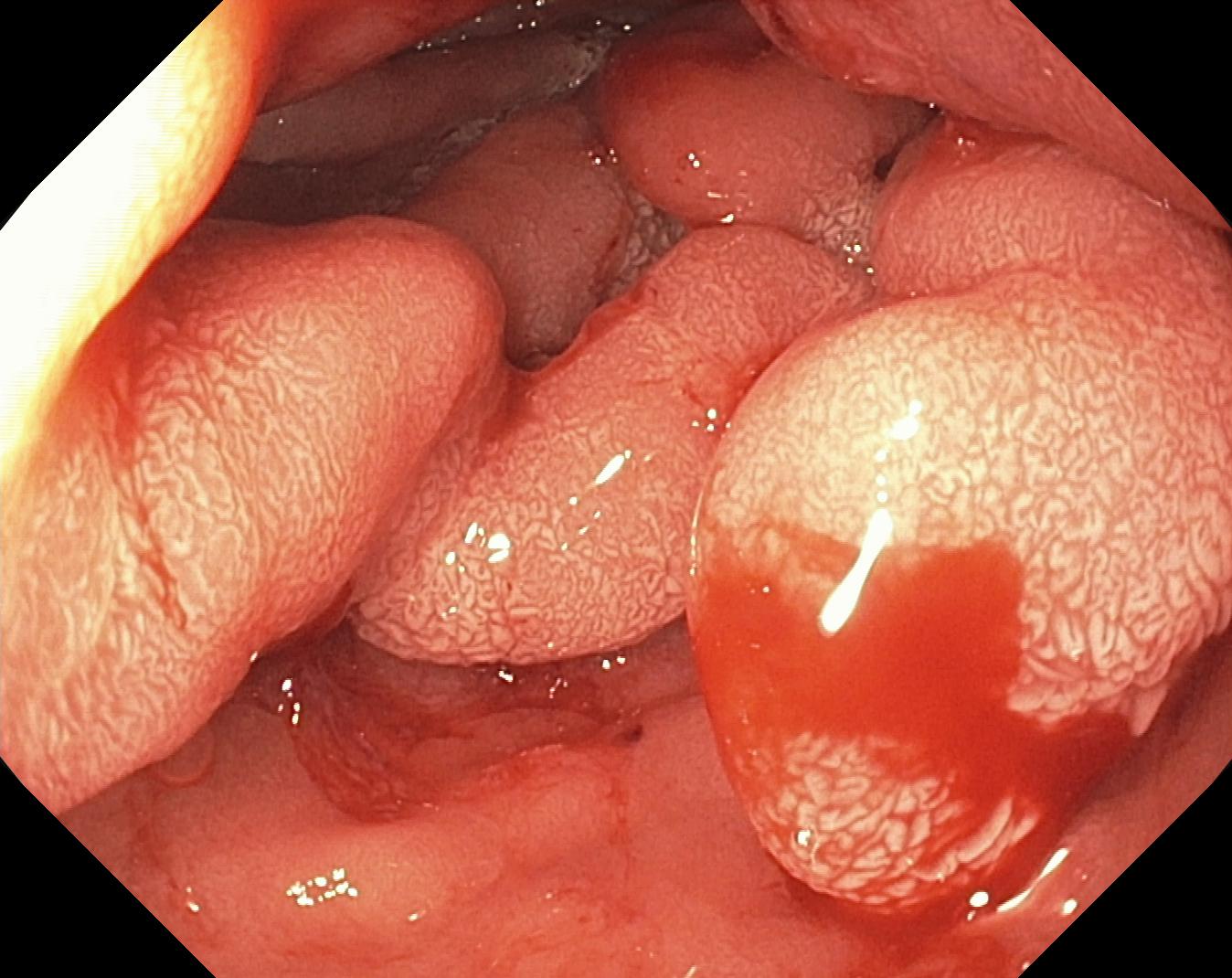}
    \includegraphics[width=1.7cm, height=1.7cm]{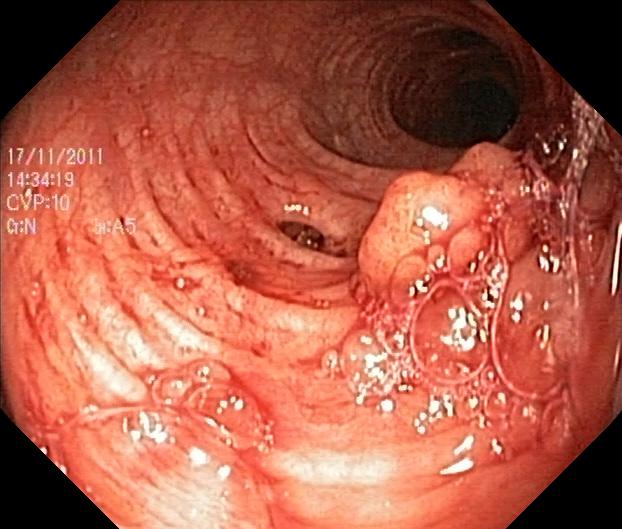}
    \includegraphics[width=1.7cm, height=1.7cm]{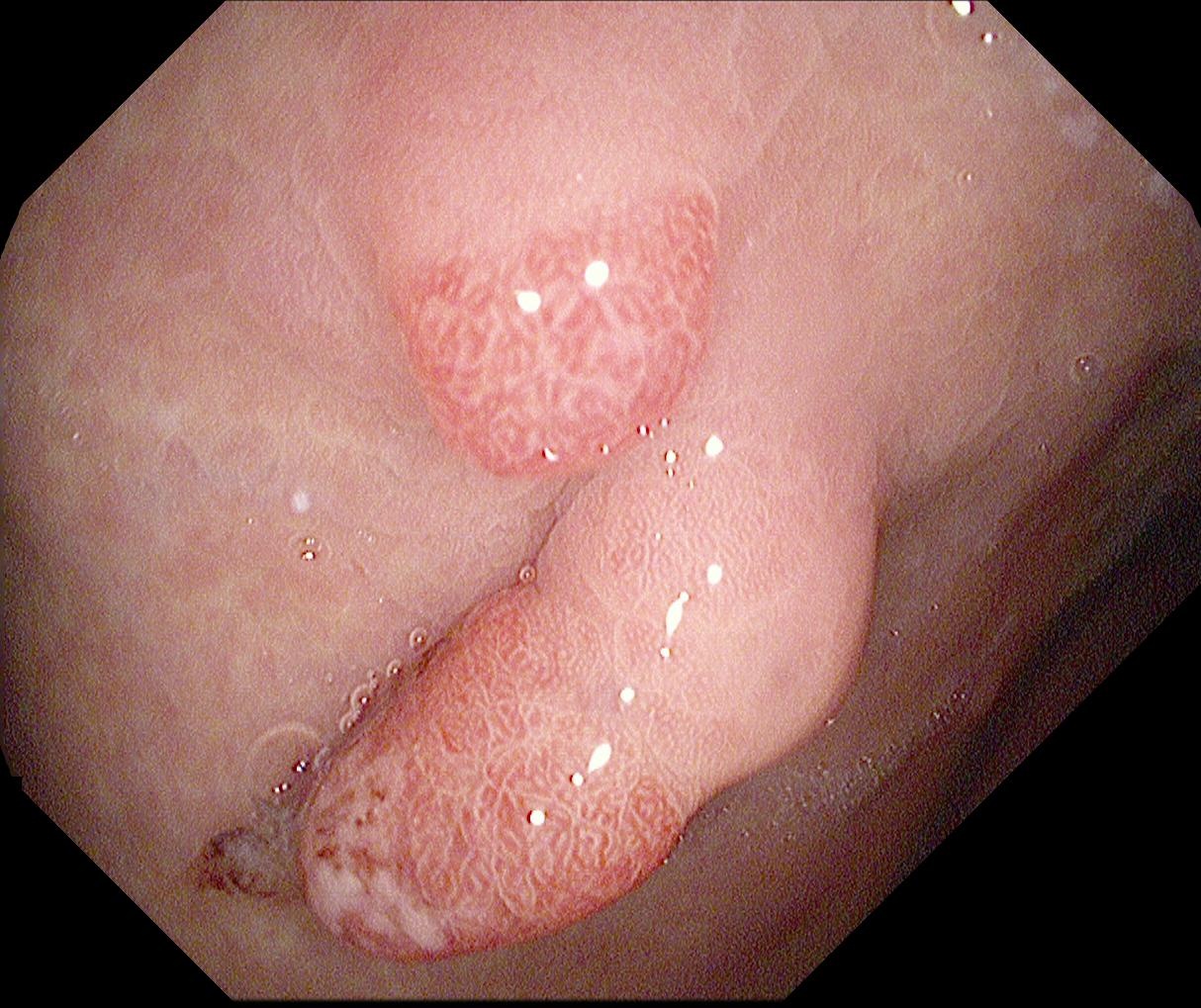}
    \includegraphics[width=1.7cm, height=1.7cm]{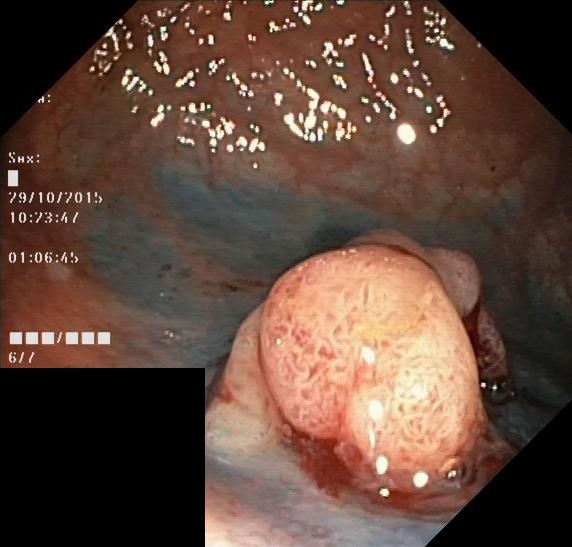}
    \caption{Example images showing the variations in shape, size, color, and appearance of polyps from the Kvasir-SEG~\cite{jha2020kvasir}.}
    \label{fig:polyp_shapes}
      \vspace{-6mm}
\end{figure}

Regardless of the achievement of colonoscopy in lowering cancer burden, the estimated adenoma miss-rate is around 6-27\%~\cite{ahn2012miss}. In a recent pooled analysis of 8 randomized tandem colonoscopy studies, polyps smaller than $10$ mm, sessile, and flat polyps~\cite{heresbach2008miss} are shown to most often be missed~\cite{zimmermann2019right}. Another reason why polyps are missed may be that the polyp either was not in the visual field or was not recognized despite being in the visual field due to fast withdrawal of the colonoscope~\cite{shaukat2015longer}. The adenoma miss-rate could be reduced by improving the quality of bowel preparation, applying optimal observation techniques, and ensuring a colonoscopy withdrawal time of at least six minutes~\cite{shaukat2015longer}. Moreover, adenoma detection rate can also be improved by using advanced techniques or devices, for example, auxiliary imaging devices, colonoscopes with increased field of view, add-on-devices, and colonoscopes with integrated inflatable, reusable balloon~\cite{matsuda2017advances}.

The structure and characteristics of a colorectal polyp changes over time at different development stages. Polyps have different shapes, sizes, colors, and appearances, which makes them challenging to analyze (see Figure~\ref{fig:polyp_shapes}). Moreover, there are challenges such as the presence of image artifacts like blurriness, surgical instruments, intestinal contents, flares, and low-quality images that can cause errors during segmentation.

Polyp segmentation is of crucial relevance in clinical applications to focus on the particular area of the potential lesion, extract detailed information, and possibly remove the polyp if necessary. A \ac{CADx} system for polyp segmentation can assist in monitoring and increasing the diagnostic ability by increasing the accuracy, precision, and reducing manual intervention. Moreover, it could lead to less segmentation errors than when conducted subjectively. Such  systems could reduce doctor's workload and improve clinical workflow. Lumen segmentation helps clinicians navigate through the colon during screening, and it can be useful to establish a quality metric for the explored colon wall~\cite{vazquez2017benchmark}. Thus, an automated \ac{CADx} system could be used as a supporting tool to reduce the miss-rate of the overlooked polyps.

A \ac{CADx} system could be used in a clinical setting if it addresses two common challenges: (i) Robustness (i.e., the ability of the model to consistently perform well on both easy and challenging images), and (ii) Generalization (i.e., a model trained on specific intervention in a specific hospital should generalize across different hospitals)~\cite{tobiasinstrument2020}. Addressing these challenges is key to designing a powerful semantic segmentation system for medical images. Generalization capability checks the usefulness of the model across different available datasets coming from different hospitals and must finally be confirmed in multi-center randomized trials. A good generalizable model could be a significant step toward developing an acceptable clinical system. A cross-dataset evaluation is crucial to check the model on the unseen polyps from other sources and test the generalizability of it. 

Toward developing a robust \ac{CADx} system, we have previously proposed \resunetplusplus~\cite{jha2019resunet++}: an initial encoder-decoder based deep-learning architecture for segmentation of medical images, which we trained, validated, and tested on the publicly available Kvasir-SEG~\cite{jha2020kvasir} and CVC-ClinicDB~\cite{bernal2015wm} datasets. In this paper, we describe how the \resunetplusplus architecture can be extended by applying \acf{CRF} and \acf{TTA} 
to further improve its prediction performance on segmented polyps.  
We have tested our approaches on six publicly available datasets, including both image datasets and video datasets. We have intentionally incorporated video datasets from colonoscopies to support the clinical significance. Usually, still-frames have at least one polyp sample. Videos have a situation where frames consist of both polyp and non-polyp. Therefore, we have tested the model on these video datasets and provided a new benchmark for the segmentation task. We have used extensive data augmentation to increase the training sample and used a comprehensive hyperparameter search to find optimal hyperparameters for the dataset. We have provided a more in-depth evaluation by including more evaluation metrics, and added justification for the \resunetplusplus, \ac{CRF}, and \ac{TTA}. 

Additionally, we have performed extensive experiments on the cross-data evaluation, in-depth analysis of best performing and worst performing cases, and comparison of the proposed method with other recent works. Moreover, we have pointed out the necessity of solving tasks related to the miss-detection of flat and sessile polyps, and showed that our combining approach could detect the overlooked polyps with high efficiency, which could be of significant importance in the clinical settings. For this, we also released a dataset consisting sessile or flat polyps publicly. Furthermore, we have emphasized the use of cross-dataset evaluation by training and testing the model with images coming from various sources to achieve the generalizability goal. 

In summary, the main contributions are as follows:

\begin{enumerate}

\item We have extended the \resunetplusplus deep-learning architecture~\cite{jha2019resunet++} for automatic polyp segmentation with \ac{CRF} and \ac{TTA} to achieve better performance. The quantitative and qualitative results shows that applying \ac{CRF}  and \ac{TTA} is effective.

\item We validate the extended architecture on a large range of datasets, i.e.,  Kvasir-SEG~\cite{jha2020kvasir}, CVC-ClinicDB~\cite{bernal2015wm}, CVC-ColonDB~\cite{bernal2012towards}, EITS-Larib~\cite{silva2014toward}, ASU-Mayo Clinic Colonoscopy Video Database~\cite{tajbakhsh2015automated} and CVC-VideoClinicDB~\cite{angermann2017towards,bernal2018polyp}, and we compare our proposed approaches with the recent \ac{SOTA} algorithm and set new a baseline. Moreover, we have compared our work with other recent works, which is often lacking in comparable studies.
\item We selected $196$ flat or sessile polyps that are usually missed during colonoscopy examination~\cite{zimmermann2019right} from the Kvasir-SEG with the help of an expert gastroenterologist. We have conducted experiments on this separate dataset to show how well our model performs on challenging polyps. Moreover, we release these polyp images and segmentation masks as a part of the Kvasir-SEG dataset so that researchers can build novel architectures and improve the results. 
\item Our model has better detection of smaller and flat or sessile polyps, which are frequently missed during colonoscopy~\cite{zimmermann2019right}, which is a major strength compared to existing works.
\item In medical clinical practice, generalizable models are essential to target patient population. Our work is focused on generalizability, previously not much explored in the community.  To promote generalizable \ac{DL} models, we have trained our models on Kvasir-SEG and CVC-ClinicDB and tested and compared the results over five publicly available diverse unseen polyp dataset. Moreover, we have mixed two diverse datasets and conducted further experiments on other unseen datasets to show the behaviour of the model on the images captured using different devices.
\end{enumerate}


\section{Related Work}
\label{sec:related}

Over the past decades, researchers have made several efforts at developing \ac{CADx} prototypes for automated polyp segmentation. Most of the prior polyp segmentation approaches were based on analyzing either the polyp's edge or its texture. More recent approaches used \ac{CNN} and pre-trained networks. Bernal et al.~\cite{bernal2015wm} introduced a novel method for polyp localization that used WM-DOVA energy maps for accurately highlighting the polyps, irrespective of its type and size. Pozdeev et al.~\cite{pozdeev2019automatic} presented a fully automated polyp segmentation framework using pixel-wise prediction based upon the \ac{FCN}. Bernal et al.~\cite{bernal2017comparative} hosted the automatic polyp detection in colonoscopy videos sub-challenge, and later on, they presented a comparative validation of different methods for automatic polyp detection and concluded that the \ac{SOTA} \ac{CNN} based methods provide the most promising results.

Akbari et al.~\cite{akbari2018polyp} used the \ac{FCN}-8S network and Otsu's thresholding method for automated colon polyp segmentation. Wang et al.~\cite{wang2018development} used the SegNet~\cite{badrinarayanan2017segnet} architecture to detect polyps. They obtained high sensitivity, specificity, and \ac{ROC} curve value. Their algorithm could achieve a speed of 25 frames per second with some latency during real-time video analysis. Guo et al.~\cite{guo2019giana} used a \ac{FCNN} model for the \ac{GIANA} polyp segmentation challenge. The proposed method won first place in the 2017 \ac{GIANA} challenge for both \ac{SD} and high definition image and won second place in the \ac{SD} image segmentation task in the 2018 \ac{GIANA} challenge. Yamada et al.~\cite{yamada2019development} developed a \ac{CADx} support system that can be used for the real-time detection of polyps reducing the number of missed abnormalities during colonoscopy.

Poorneshwaran et al.~\cite{poomeshwaran2019polyp} used a Generative Adversarial Network (GAN) for polyp image segmentation. Kang et al.~\cite{kang2019ensemble} used Mask R-CNN, which relies on  ResNet50 and ResNet101, as a backbone structure for automatic polyp detection and segmentation. Ali et al.~\cite{ali2019endoscopy} presented various detection and segmentation methods that could classify, segment, and localize artifacts. Additionally, there are several recent really interesting studies on polyp segmentation~\cite{nguyen2019robust,de2019training,sun2019colorectal,jha2020doubleu}. They are useful steps toward building an automated polyp segmentation system. There are also some works which have hypothesized that coupling the existing architecture by applying careful post-processing technique could improve the model performance~\cite{jha2019resunet++, ibtehaz2020multiresunet}. 

From the presented related work, we observe that automatic \ac{CADx} systems in the area of polyp segmentation are becoming mature. Researchers are conducting a variety of studies with different designs ranging from a retrospective study, prospective study, to post hoc examination of the prospectively obtained dataset.  Some of the models achieve very high performance with smaller training and test datasets~\cite{brandao2018towards,wang2018development,jha2019resunet++}. The algorithms used for building the models are the ones that use handcrafted-, \ac{CNN}- or pre-trained-features from ImageNet~\cite{deng2009imagenet}, where \ac{DL} based algorithms are outperforming and gradually replacing the traditional handcrafted or \ac{ML} approaches.  Additionally, the performance of the models improves by the use of advance \ac{DL} algorithms, especially designed for polyp segmentation task or any other similar biomedical image segmentation task. Moreover, there is interest for testing the proposed architectures with more than one dataset\cite{wang2018development,jha2019resunet++}.

The main drawbacks in the field are the minimal effort applied towards testing the generalizability of the \ac{CADx} system possible to achieve with the cross-dataset test. Additionally, there is almost no effort involved in designing an universal model that could accurately segment polyp coming from different sources,  critical for the development of \ac{CADx} for automated polyp segmentation. Besides, most of the current works have proposed algorithms that are tested on single, often small, imbalanced, and explicitly handpicked datasets. This renders conclusions regarding the performance of the algorithms almost useless (compared to other areas in \ac{ML} like, for example, natural image classification or action recognition where the common practice is to test on more than one dataset and make source code and datasets publicly available). Additionally, the used datasets are often not public available (restricted and difficult to access), and the total number of images and videos used in the study are not sufficient to believe that the system is robust and generalizable for use in clinical trials. For instance, the model can produce output segmentation map with high sensitivity and precision on a particular dataset and completely fails on other modality images. Moreover, existing work often use small training and test datasets. These current limitations make it harder to develop a robust and generalizable systems.

Therefore, we aim to develop a \ac{CADx} based support system that could achieve high performances irrespective of the datasets. To achieve the goal, we have done extensive experiments on various colonoscopy images and video datasets. Additionally, we have mixed the dataset from multiple centers and tested it on other diverse unseen datasets to achieve the goal of building a generalizable and robust \ac{CADx} system that produces no segmentation errors.  Moreover, we set a new benchmark for the publicly available datasets that can be improved in the future. 
 \begin{figure}[t!]
   \centering
   \scalebox{0.8}{
  \includegraphics[trim=1.3cm 2cm 2.5cm 1.1cm, clip, width=\linewidth]{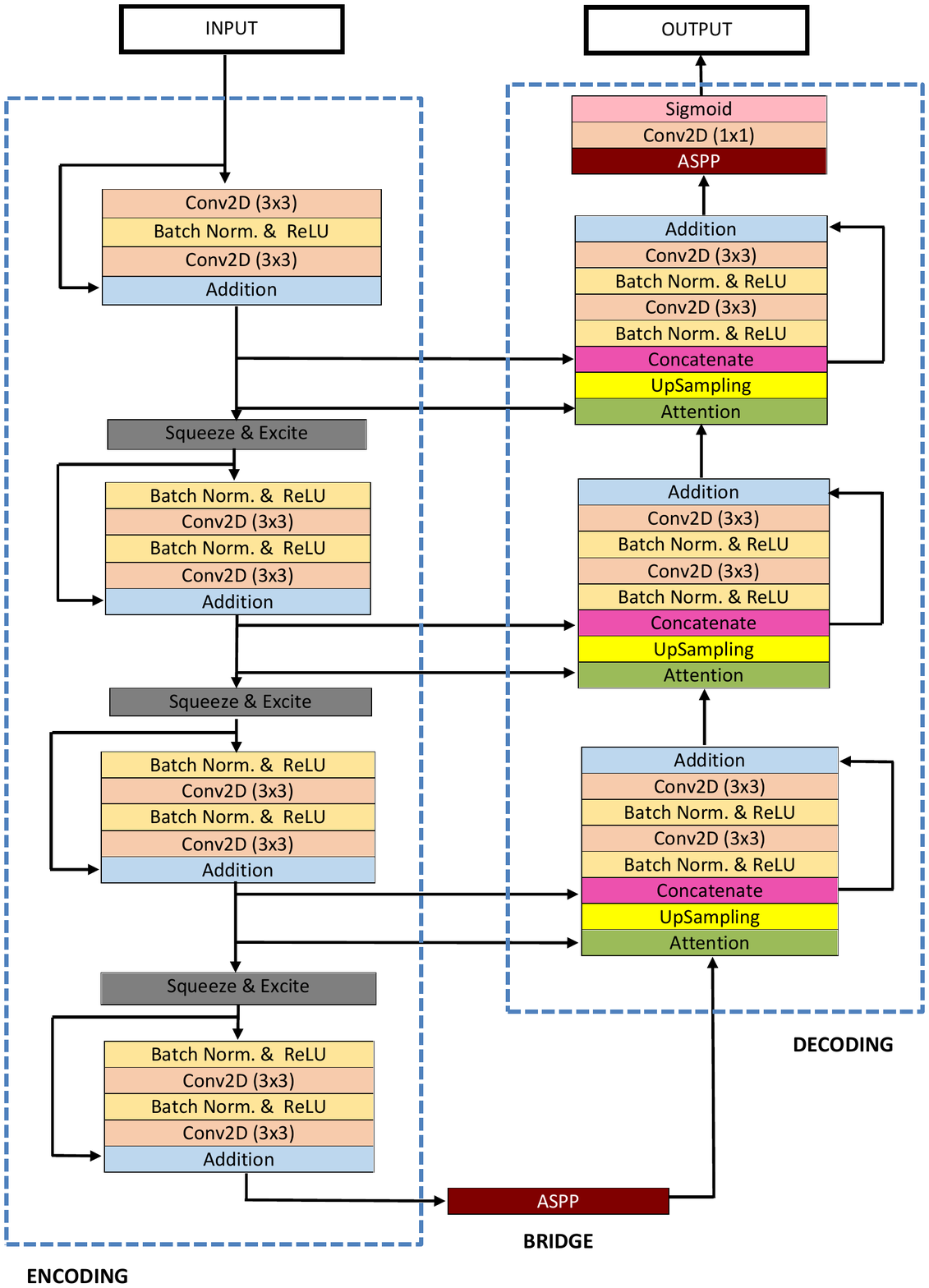}}
  \vspace{-3mm}
 \caption{\resunetplusplus architecture~\cite{jha2019resunet++}}
  \label{fig:proposed_method}
  \vspace{-3.1mm}
 \end{figure}
\section{The \resunetplusplus Architecture}
\label{proposed_archicture}

\resunetplusplus is a semantic segmentation deep neural network designed for medical image segmentation. The backbone for \resunetplusplus architecture is ResUNet~\cite{zhang2018road}: an encoder-decoder network and based on U-Net~\cite{ronneberger2015u}. The proposed architecture takes the benefit of residual block, squeeze and excite block~\cite{hu2018squeeze}, \ac{ASPP}~\cite{chen2017rethinking}, and attention block~\cite{vaswani2017attention}. What distinguishes \resunetplusplus from ResUNet is the use of squeeze-and-excitation blocks (marked in dark gray) at the encoder, the \ac{ASPP} block, (marked in the dark red) at bridge and decoder, and the attention block (marked in light green) at the decoder (see Figure~\ref{fig:proposed_method}). 

In the \resunetplusplus model, we introduce the sequence of squeeze and excitation block to the encoder part of the network. Additionally, we replace the bridge of ResUNet with \ac{ASPP}. In the decoder stage, we introduce a sequence of attention block, nearest-neighbor up-sampling, and concatenate it with the relevant feature map from the residual block of the encoder through skip connection. This process is followed by the residual unit with identity mapping, as shown in Figure~\ref{fig:proposed_method}. 

We also introduce a series of additional skip connections from the residual unit of the encoder section to the attention block of the decoder section. We assign the number of filters $[32, 64, 128, 256, 512]$, along with the levels in the encoder section, which are the values in our \resunetplusplus architecture. These filter combinations achieved the best results in our \resunetplusplus experiment. In the decoder section, the number of the filter is reversed, and the sequence becomes $[512, 256, 128, 64, 32]$. As the semantic gap between the feature map of the encoder and decoder blocks are supposed to decrease, the number of filters in the convolution layers of the decoder block are also decreased to achieve better semantic coverage. Through this, we ensure that the overall quality of the feature maps is more alike to the ground truth mask. This is especially important as the loss in semantic space is likely to decrease, and therefore it will become more feasible to find a meaningful representation in semantic space.

The overall \resunetplusplus architecture consists of one stem block with three encoder blocks, an \ac{ASPP} between the encoder and the decoder, and three decoder blocks. All the encoder and decoder blocks use the standard residual learning approach. Skip connections are introduced between encoder and decoder for the propagation of information. The output of the last decoder block is passed through the \ac{ASPP}, followed by a $1\times1$ convolution and a sigmoid activation function. All convolutional layers except for the output layer are batch normalized~\cite{ioffe2015batch} and are activated by a ~\ac{ReLU} activation function~\cite{lecun2015deep}. Finally, we get the output as binary segmentation maps. A brief explanation of each block is provided in the following sub-sections.

\subsection{Residual Blocks}
Training a deep neural network by expanding network depth can potentially improve overall performance. Nevertheless, simply stacking the \ac{CNN} layer could also hamper the training process and cause exploding/vanishing gradient when backpropagation occurs~\cite{he2016deep}. Residual connections facilitate the training process by directly routing the input information to the output and preserves the nobility of the gradient flow. The residual function simplifies the objective of optimization without any additional parameters and boosts the performance, which is the inspiration behind the deeper residual-based network~\cite{wang2019nested}. Equation~\eqref{equation1} below shows the working principle.
\begin{align}
\label{equation1}
 y_n = {F(x_n,W_n) + x_n}
\end{align}
Here, $x_n$ is the input and $F(\cdot)$ is the residual function.  
 The residual units consist of numerous combinations of~\ac{BN},~\ac{ReLU}, and convolution layers.  A detailed description of the combinations used and their impact can be found in the work of He et al.~\cite{he2016identity}. We have employed the concept of a pre-activation residual unit in the \resunetplusplus architecture from ResUNet. 
 
\subsection{Squeeze and Excitation block}
The~\ac{SE} block is the building block for the \ac{CNN} that re-calibrates channel-wise feature response by explicitly modeling interdependencies between the channels~\cite{hu2018squeeze}. The~\ac{SE} block learns the channel weights through global spatial information that increases the sensitivity of the effective feature maps, whereas it suppresses the irrelevant feature maps~\cite{jha2019resunet++}.  The feature maps produced by the convolution have only access to the local information, meaning they have no access to the global information left by the local receptive field. To address this limitation, we perform a squeeze operation on the feature maps using the global average pooling to generate a global representation. We then use the global representation and perform sigmoid activation that helps us to learn a non-linear interaction between the channels, and capture the channel-wise dependencies. Here, the sigmoid activation output acts as a simple gating mechanism that ensures us to adaptively recalibrate the feature maps produced by the convolution. The adaptive recalibration or excitation operation explicitly models the interdependencies between the feature channels. The \ac{SE} net has the capability of generalizing exceptionally well across various datasets~\cite{hu2018squeeze}. In the \resunetplusplus architecture, we have stacked the \ac{SE} block together with the residual block for improving the performance of the network, increasing the effective generalization across different medical datasets.
 
\subsection{Atrous Spatial Pyramidal Pooling}
Since the introduction of Atrous convolution by Chen et al.~\cite{chen2014semantic} to control the field-of-view to capture contextual information at multi-scale precisely, it has shown promising results for semantic image segmentation. Later, Chen et al.~\cite{chen2017deeplab} proposed \ac{ASPP}, which is a parallel atrous convolution block to capture multiple-scale information simultaneously. \ac{ASPP} captures the contextual information at different scales, and multiple parallel atrous convolutions with varying rates in the input feature map are fused~\cite{chen2017deeplab}. In \resunetplusplus, we use \ac{ASPP} as a  bridge between the encoder and the decoder sections, and after the final decoder block. We adopt \ac{ASPP}  in \resunetplusplus to capture the useful multi-scale information between the encoder and the decoder. 
\subsection{Attention Units}
Chen et al.~\cite{chen2016attention} proposed an attention model that can segment natural images by multi-scale input processing. Attention model is an improvement over average and max-pooling baseline and allows to visualize the features importance at different scales and positions~\cite{chen2016attention}. With the success of attention mechanisms, various medical image segmentation methods have integrated an attention mechanism into their architecture~\cite{wang2018deep,jha2019resunet++,nie2018asdnet,sinha2019multi}. The attention block gives importance to the subset of the network to highlight the most relevant information. We believe that the attention mechanism in our architecture will boost the effectiveness of the feature maps of the network by capturing the relevant semantic class and filtering out irrelevant information. Motivated by the recent achievement of attention mechanism in the field of medical image segmentation and computer vision in general, we have integrated an attention block at the decoder part of the \resunetplusplus model. 

\subsection{Conditional Random Field}
\acf{CRF} is a popular statistical modeling method used when the class labels for different inputs are not independent (e.g., image segmentation tasks). \ac{CRF} can model useful geometric characteristics like shape, region connectivity, and contextual information~\cite{alam2018conditional}. Therefore, the use of \ac{CRF} can further improve the models capability to capture contextual information of the polyps and thus improve overall results. We have used \ac{CRF} as a further step to produce more refined output to the test dataset for improving the segmentation results. we have used a dense \ac{CRF} for our experiment.
\subsection{Test Time Augmentation}
\acf{TTA} is a technique of performing reasonable modifications to the test dataset to improve the overall prediction performance. In \ac{TTA}, augmentation is applied to each test image, and multiple augmented images are created. After that, we make predictions on these augmented images, and the average prediction of each augmented image is taken as the final output prediction. Inspired by the improvement of recent~\ac{SOTA}~\cite{guo2019giana}, we have used \ac{TTA} in our work. In this paper, we utilize both horizontal and vertical flip for \ac{TTA}. 

\section{Experiments}
\label{sec:material}


\begin{figure}
 \centering
        \includegraphics[width=1.7cm, height=1.7cm]{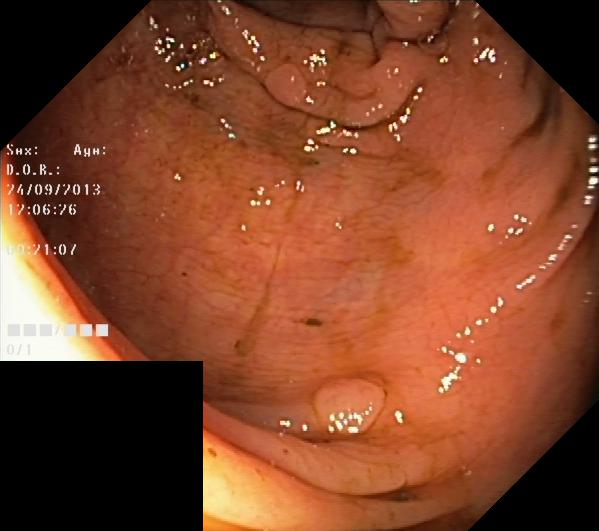}
        \includegraphics[width=1.7cm, height=1.7cm]{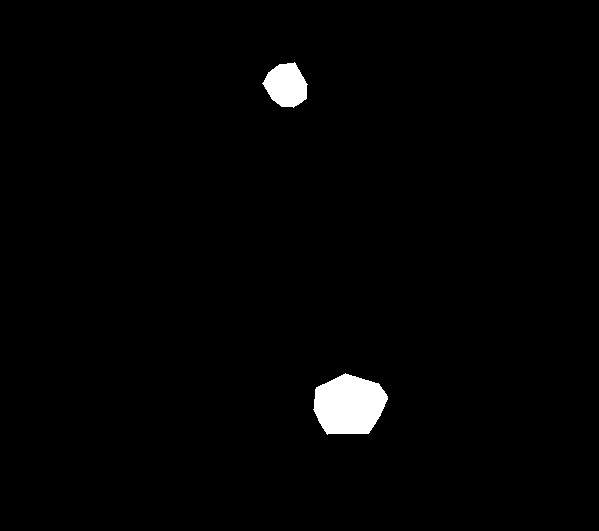}
        \includegraphics[width=1.7cm, height=1.7cm]{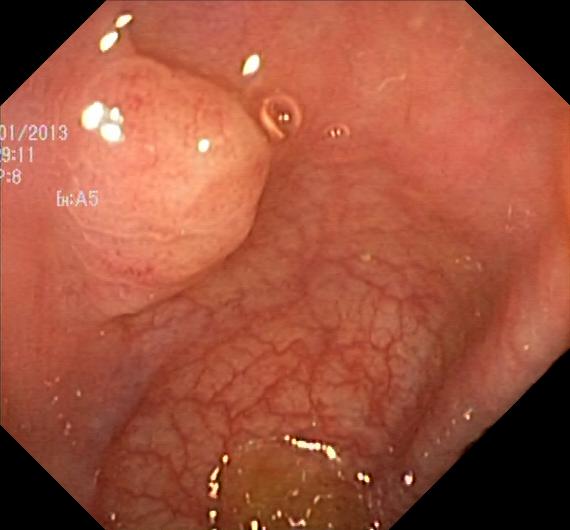}
        \includegraphics[width=1.7cm, height=1.7cm]{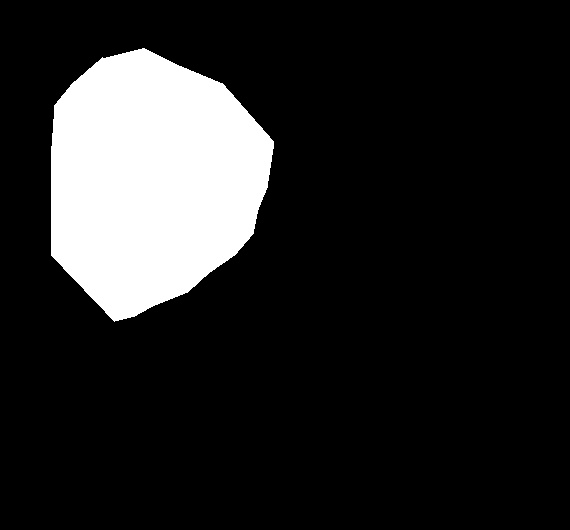} \\
        \vspace{1mm}
        \includegraphics[width=1.7cm, height=1.7cm]{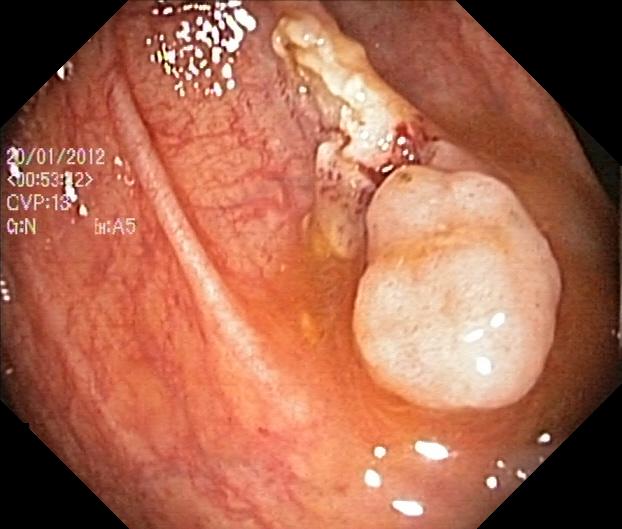}
        \includegraphics[width=1.7cm, height=1.7cm]{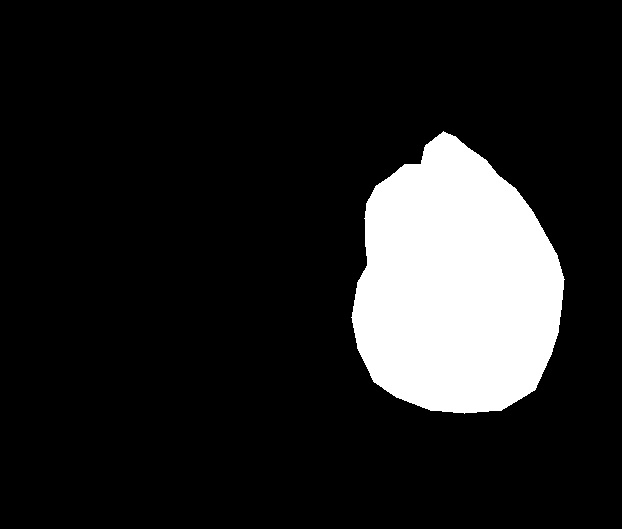}
        \includegraphics[width=1.7cm, height=1.7cm]{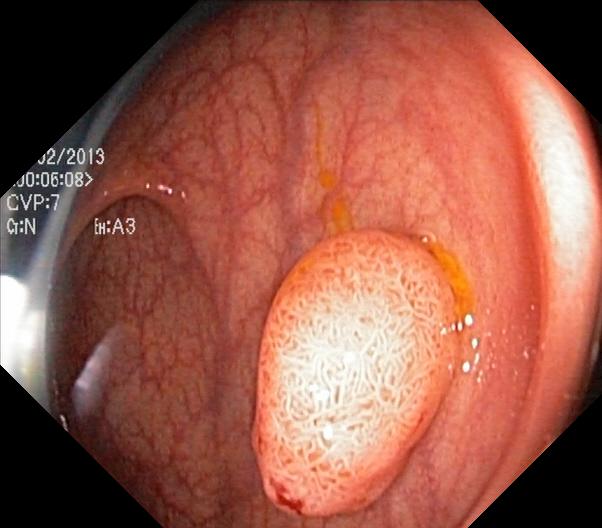}
        \includegraphics[width=1.7cm, height=1.7cm]{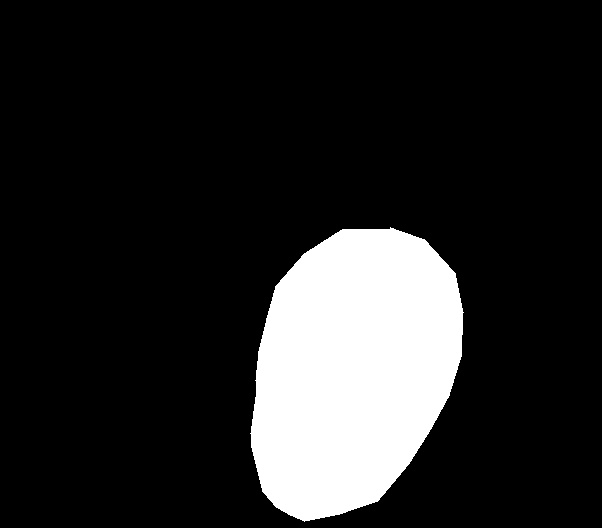}
    
    \caption{Example polyp and corresponding ground truth from the Kvasir-SEG}
    \label{fig:polyp_with_mask}
    \vspace{-3mm}
\end{figure}

\subsection{Datasets}
\begin{table} [t!]

 \caption{The biomedical segmentation datasets used in our experiments}
    \label{table:datasettable}
    \scriptsize
    \centering
          \begin{tabular}{ l c c c c} 
                \toprule
                Dataset & \shortstack{Images} & Input size & Availability\\ 
              \bottomrule
                Kvasir-SEG~\cite{jha2020kvasir}  & 1000 & Variable & Public\\ 
                 CVC-ClinicDB~\cite{bernal2015wm} & 612 & $384\times 288$ & Public \\ 
                 CVC-ColonDB~\cite{bernal2012towards} & 380  & $574\times 500$ & Public\\ 
                 ETIS Larib Polyp DB~\cite{silva2014toward}& 196 & $1225 \times 966$ &Public \\ 
                CVC-VideoClinicDB~\cite{angermann2017towards,bernal2018polyp}$^\dag$$^\diamond$ &11,954 & $384\times 288$ & Public \\ 
                ASU-Mayo Clinic dataset~\cite{tajbakhsh2015automated}$^\dag$&18,781&$688 \times 550$ & Copyrighted\\ 
                Kvasir-Sessile$^\bullet$ & 196 & Variable & Public\\ 
                 \bottomrule
                \multicolumn{4}{l}{$^\dag$ Ground truth for test data not available} \hspace{.1cm} $^\diamond${Ground truth oval or circle shaped}\\  
                \multicolumn{4}{l}{$^\bullet$ Part of Kvasir-SEG\cite{jha2020kvasir}, only sessile polyps} \\  
                
                 \vspace{-9mm}
\end{tabular}
\end{table}	

We have used six different datasets of segmented polyps with ground truths in our experiments as shown in Table~\ref{table:datasettable}, i.e., Kvasir-SEG~\cite{jha2020kvasir}, CVC-ClinicDB~\cite{bernal2015wm}, CVC-ColonDB~\cite{bernal2012towards}, ETIS Larib Polyp DB~\cite{silva2014toward}, CVC-VideoClinicDB~\cite{angermann2017towards,bernal2018polyp} and ASU-Mayo Clinic dataset~\cite{tajbakhsh2015automated}. They vary e.g., regarding number of images, image resolution, availability, devices used for capturing and the accuracy of the segmentation masks. One example is given from the Kvasir-SEG in Figure~\ref{fig:polyp_with_mask}. The Kvasir-SEG dataset includes 196 polyps smaller than 10 mm classified as Paris class 1 sessile or Paris class IIa. We have released this dataset seperately as subset of Kvasir-SEG.  Note that for CVC-VideoClinicDB, we have only used the data from the CVC-VideoClinicDBtrainvalid folder since only these data have ground truth masks. Moreover, the ASU-Mayo Clinic dataset, which was made available at the ``Automatic Polyp Detection in Colonoscopy Videos" sub-challenge at Endovis 2015
 had ten normal videos (negative shots) and ten videos with polyps. However, the test subset is not available because of issues related to licensing.  In our experiment, while training, validating, testing with 80:10:10 split on the ASU-Mayo, we used all 20 videos for experimentation. However, for the cross-dataset test (i.e., Tables \ref{table:crossdatakvasir} and \ref{table:crossdataCVCclinicDB}), we only tested on ten positive polyp videos.

\subsection{Evaluation Method}


To evaluate polyp segmentation methods, where individual pixels should be identified and marked, we use metrics used in earlier research~\cite{bernal2017comparative,guo2019giana,wang2018development,ali2019endoscopy,jha2020kvasir,pogorelov2017kvasir} and in competitions like GIANA\footnote{https://giana.grand-challenge.org/}, comparing the correctly and wrongly identified pixels of findings. The \ac{DSC} and the \ac{IOU} are the most commonly used metrics. We use the DSC to compare the similarity between the produced segmentation results and the original ground truth. Similarly, the \ac{IOU} is used to compare the overlap between the output mask and original ground truth mask of the polyp. The \ac{mIoU} calculates \ac{IOU} of each semantic class of the image and compute the mean over all the classes. There is a correlation between \ac{DSC} and \ac{mIoU}. However, we calculate both the metrics to provide a comprehensive results analysis that could lead to better understanding of the results.

Moreover, other often-used metrics for the binary classification are recall (true positive rate) and precision (positive predictive value). For the polyp segmentation, precision is the ratio of the number of correctly segmented pixels versus the total number of all the pixels. Similarly, recall is the ratio of correctly segmented pixel versus the total number of pixels present in the ground truth. In the polyp image segmentation, precision and recall are used to indicate over-segmentation and under-segmentation. For formal definitions and formulas, see the definitions in for example~\cite{jha2020kvasir,pogorelov2017kvasir}. Finally, the \acf{ROC} curve analysis is also an important metric to characterize the performance of the binary classification system. In our study, we therefore calculate \ac{DSC}, \ac{mIoU}, recall, precision, and \ac{ROC} when evaluating the segmentation models.

\subsection{Data Augmentation}
Data augmentation is a crucial step in increasing the number of polyp samples. This solves the data insufficiency problem, improves the performance of the model, and help to reduce over-fitting. We have used a large number of different data augmentation techniques to increase the training sample. We divide all the polyp datasets into training, validation, and testing sets using the ratio of 80:10:10 based on the random distribution except for the mixed datasets. After splitting the dataset, we apply data augmentation techniques such as center crop, random rotation, transpose, elastic transform, grid distortion, optical distortion, vertical flip, horizontal flip, grayscale, random brightness, random contrast, hue saturation value, RBG shift, course dropout, and different types of blur. For cropping the images, we have used a crop size of $256 \times 256$ pixels. For the experiment, we have resized the complete training, validation, and testing dataset to $256 \times 256$ pixels to reduce the computational complexity. We have only augmented the training dataset. The validation data is not augmented, and the test datasets were augmented while evaluation using \ac{TTA}.
\subsection{Implementation and Hardware Details}
We have implemented all the models using the Keras framework~\cite{chollet2015keras} with Tensorflow~\cite{abadi2016tensorflow} as a backend. Source code of our implementation and information about our experimental setup are made publicly available on Github\footnote{\url{https://github.com/DebeshJha/ResUNet-with-CRF-and-TTA}}. Our experiments were performed using a Volta 100 Tensor Core GPU on a Nvidia DGX-2 AI system capable of 2-petaFLOPS tensor performance. We used a Ubuntu 18.04.3LTS operating system with Cuda 10.1.243 version installed. We have performed different experiments with different sets of hyperparameters manually on the same dataset in order to select the optimal set of hyperparameters for the \resunetplusplus. Our model performed well with the batch size of $16$, Nadam as an optimizer, binary cross-entropy as the loss function, and learning rate of $1\mathrm{e}{-5}$. The dice loss function was also competitive. These hyperparameters were chosen based on the empirical evaluation. All the models were trained for $300$ epochs. We have used early stopping to prevent the model from over-fitting. To further improve the results, we have used stochastic gradient descent with warm restarts (SGDR). All the hyperparameters were same except the learning rate, which was adjusted based on the requirement. We have also included the Tensorboard for the analysis and visualization of the results. 

\section{Results}
\label{sec:results}
In our previous work, we have showed that \resunetplusplus outperforms the \ac{SOTA} UNet~\cite{ronneberger2015u} and ResUNet~\cite{zhang2018road} models trained on Kvasir-SEG and CVC-ClinicDB dataset\cite{jha2019resunet++}. In this work, we aim to improve the results of \resunetplusplus by utilizing further hyperparameter optimization, \ac{CRF} and \ac{TTA}. In this section, we present and compare the results of \resunetplusplus with \ac{CRF}, \ac{TTA}, and both approaches combined on the same dataset, mixed dataset, and cross-dataset. Although a direct comparison of approaches from the literature is difficult due to different testing mechanisms used by various authors, we nonetheless compare the results with the recent work for the evaluation.


\begin{table}[!t]
 \caption{Results comparison on Kvasir-SEG}
    \label{table:resultkvasir}
\scriptsize
    \centering
           \begin{tabular}{ l c c c c} 
                \toprule
                Method & DSC & mIoU & Recall & Precision\\ 
              \bottomrule
                 UNet~\cite{ronneberger2015u} & 0.7147 & 0.4334 & 0.6306 & \textbf{0.9222} \\
                 ResUNet~\cite{zhang2018road} & 0.5144 & 0.4364  & 0.5041 & 0.7292 \\ 
                 ResUNet-mod~\cite{zhang2018road} & 0.7909 & 0.4287 & 0.6909 & 0.8713 \\ 
                 \resunetplusplus~\cite{jha2019resunet++} & 0.8119  & 0.8068 & 0.8578 & 0.7742\\ 
                 \resunetplusplus + CRF & 0.8129 & 0.8080 & 0.8574 & 0.7775\\ 
                 \resunetplusplus TTA & 0.8496 & 0.8318 & \textbf{0.8760} & 0.8203\\
                 \resunetplusplus +TTA + CRF  & \textbf{0.8508} & \textbf{0.8329} & 0.8756 &0.8228\\
               \bottomrule
\end{tabular}
     \vspace{-3mm} 
\end{table}						

\subsection{Results comparison on Kvasir-SEG dataset}
\begin{figure*} [t!]
    \centering
    \includegraphics [width=0.8\textwidth ]{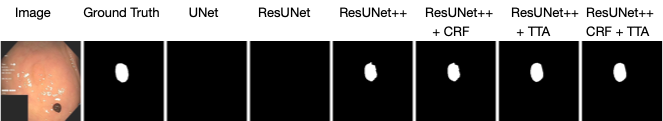}\vspace{0.5mm}\\
    \includegraphics [width=0.8\textwidth ]{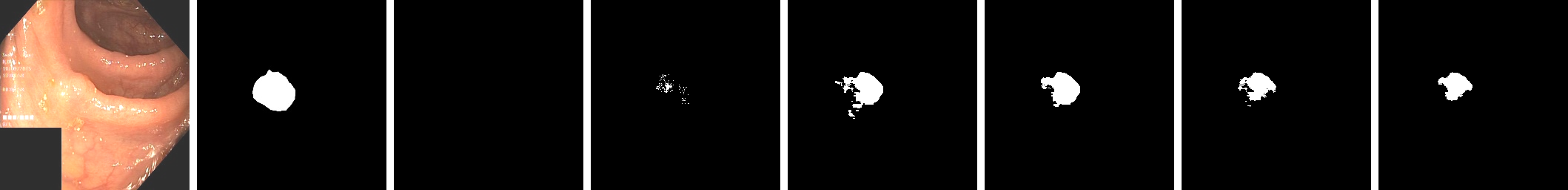}\vspace{0.5mm}\\
    \includegraphics [width=0.8\textwidth ]{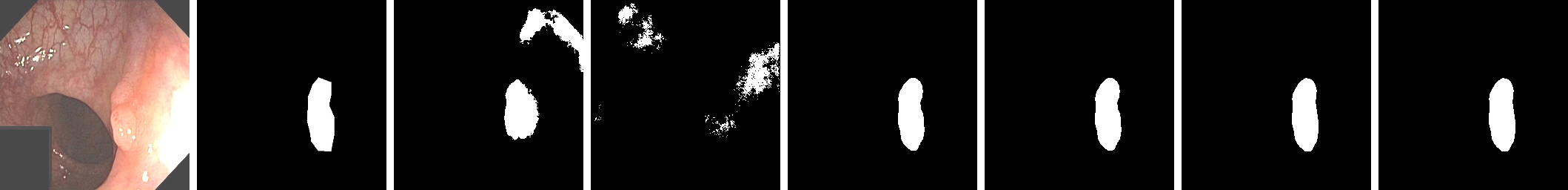}\vspace{0.5mm}\\
    \includegraphics [width=0.8\textwidth ]{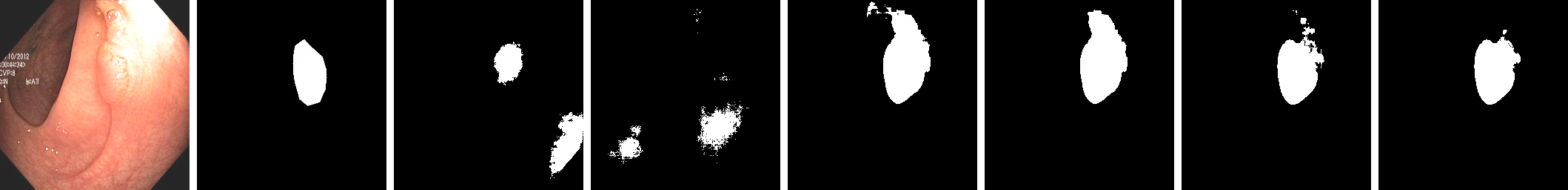}\vspace{0.5mm}\\
    \caption{Qualitative results comparison of the proposed models with UNet, ResUNet, and \resunetplusplus. The figure shows the example of polyps that are usually missed-out during colonoscopy examination. We see that there is a high similarity between ground truth and predicted mask for the proposed models.} 
  \label{fig:flatpolypkvasir}
  \vspace{-3mm}
\end{figure*}

Table~\ref{table:resultkvasir} and Figure~\ref{fig:flatpolypkvasir} show the quantitative and qualitative results comparison. Figure~\ref{fig:figure6} shows the \ac{ROC} curve for all the models. As seen in the quantitative results
(Table~\ref{table:resultkvasir}), qualitative results (Figure~\ref{fig:flatpolypkvasir}), and \ac{ROC} curve (Figure~\ref{fig:figure6}), our proposed methods outperform ResUNet++ on the Kvasir-SEG dataset. The improvement in results demonstrates the advantage of the use of the \ac{TTA}, \ac{CRF} and their combinations.

\begin{figure}[t!]
    \centering
    \includegraphics [width=9cm]{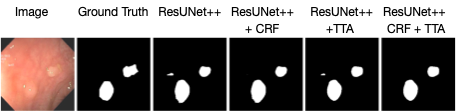}\vspace{0.5mm}\\
    \includegraphics [width=9cm]{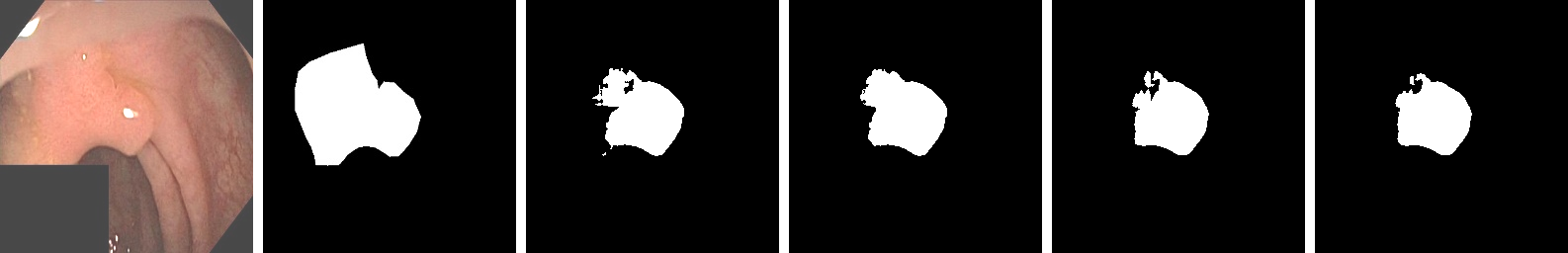}\vspace{0.5mm}\\
    \includegraphics [width=9cm]{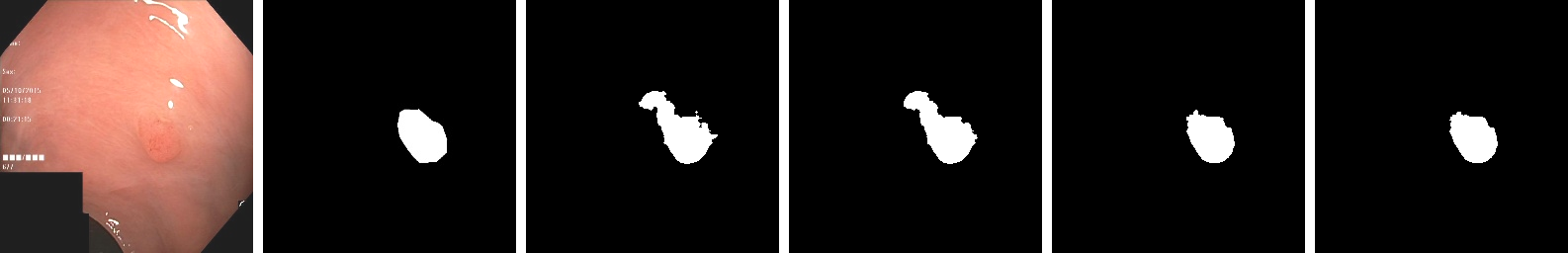}\vspace{0.5mm}\\
    \includegraphics [width=9cm]{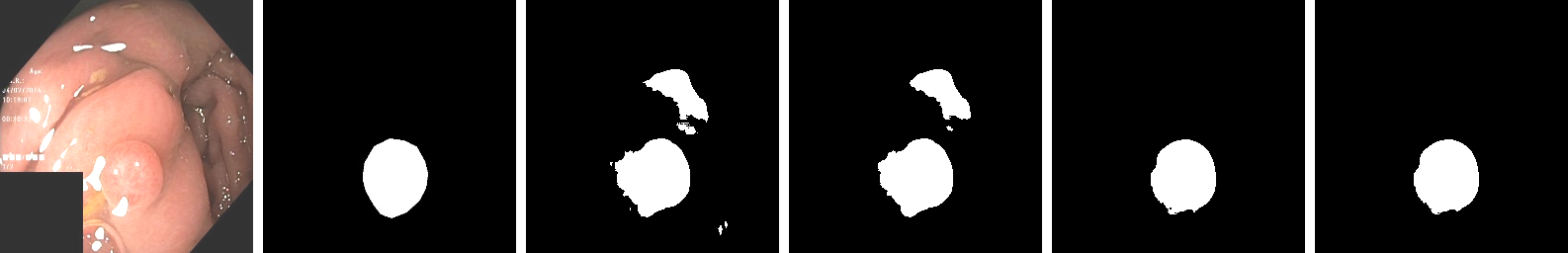}\vspace{0.5mm}\\
    \caption {Result of model trained on CVC-ClinicDB and tested on Kvasir-SEG}
  \label{fig:cross-data612kvasir}
    \vspace{-3mm}
\end{figure}

\begin{figure}[t!]
    \centering
    \includegraphics [width=9cm]{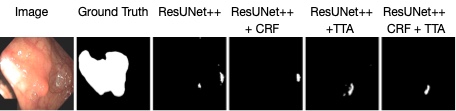}\vspace{0.5mm}\\
    \includegraphics [width=9cm]{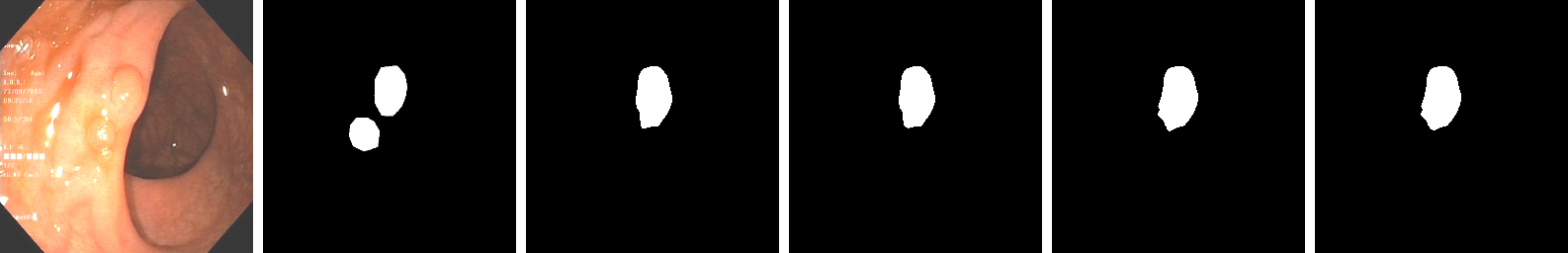}\vspace{0.5mm}\\
    \includegraphics [width=9cm]{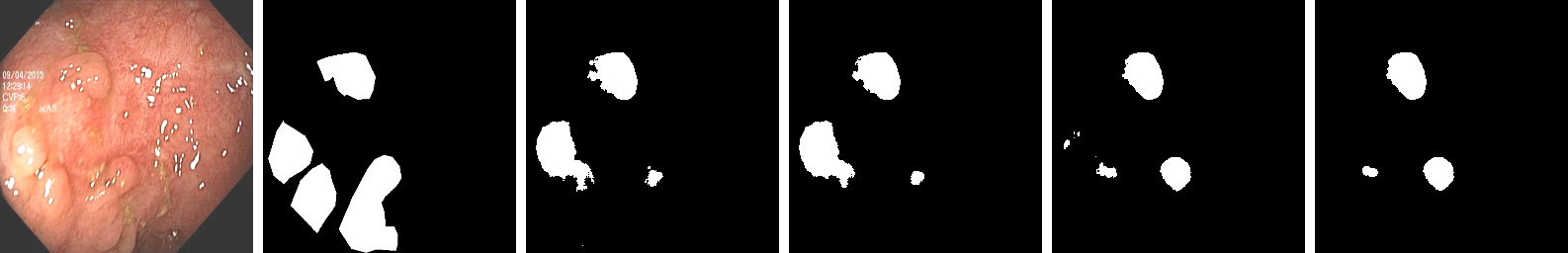}\vspace{0.5mm}\\
    \includegraphics [width=9cm]{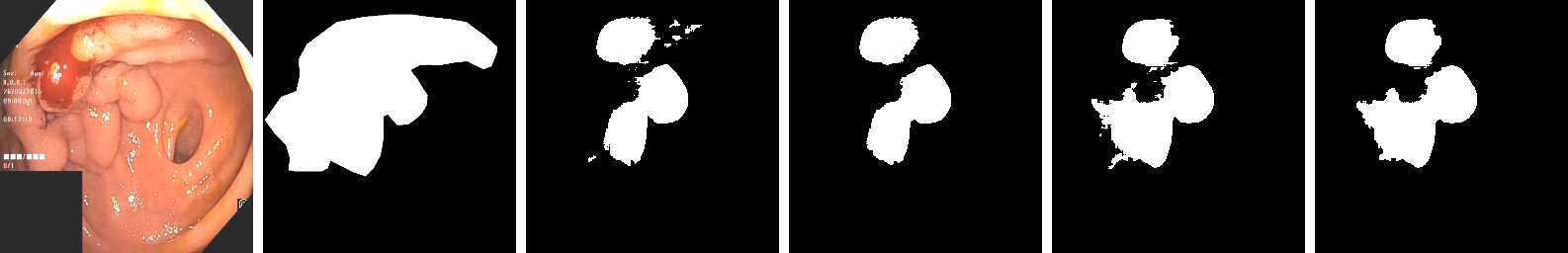}\vspace{0.5mm}\\
   \caption{Example images where the proposed models fails on Kvasir-SEG}
  \label{fig:samekvasirfail}
  \vspace{-0mm}
\end{figure}


\begin{figure}[t!]
    \centering
    \scalebox{0.8}{
     \includegraphics [width=.7\linewidth]{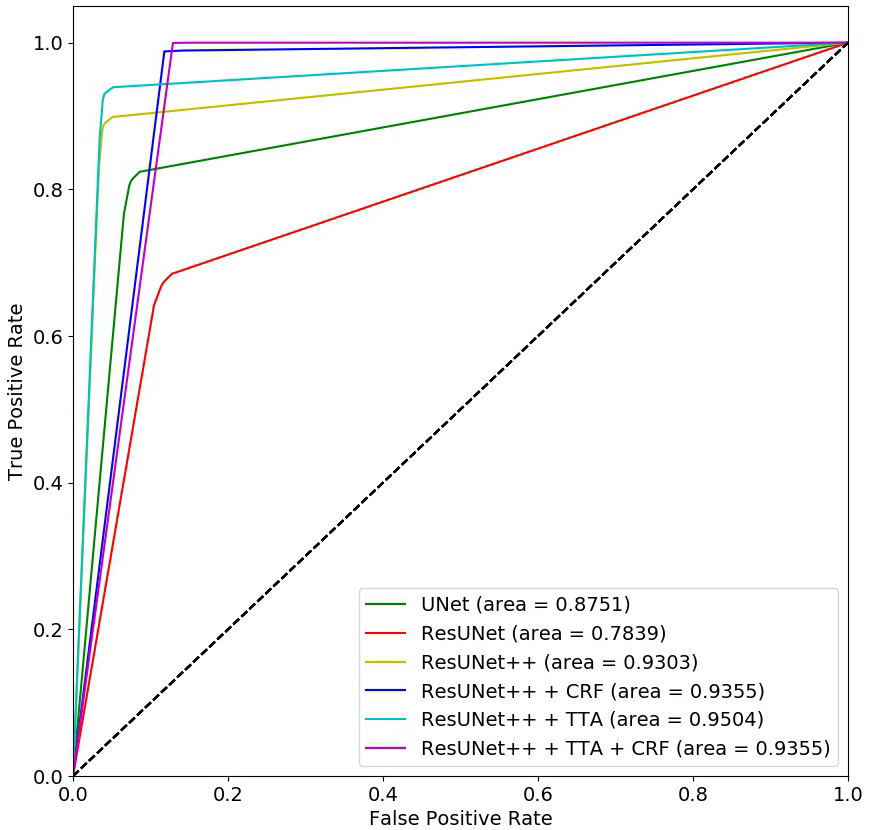}}
    \caption{ROC curve of proposed models on the Kvasir-SEG}
    \label{fig:figure6}
\end{figure}
\begin{figure}[t!]
    \vspace{-4mm} 
    \centering
    \scalebox{0.8}{
    \includegraphics [trim=1.9cm 1.2cm 1.9cm 2.5cm, clip,width=.9\linewidth] {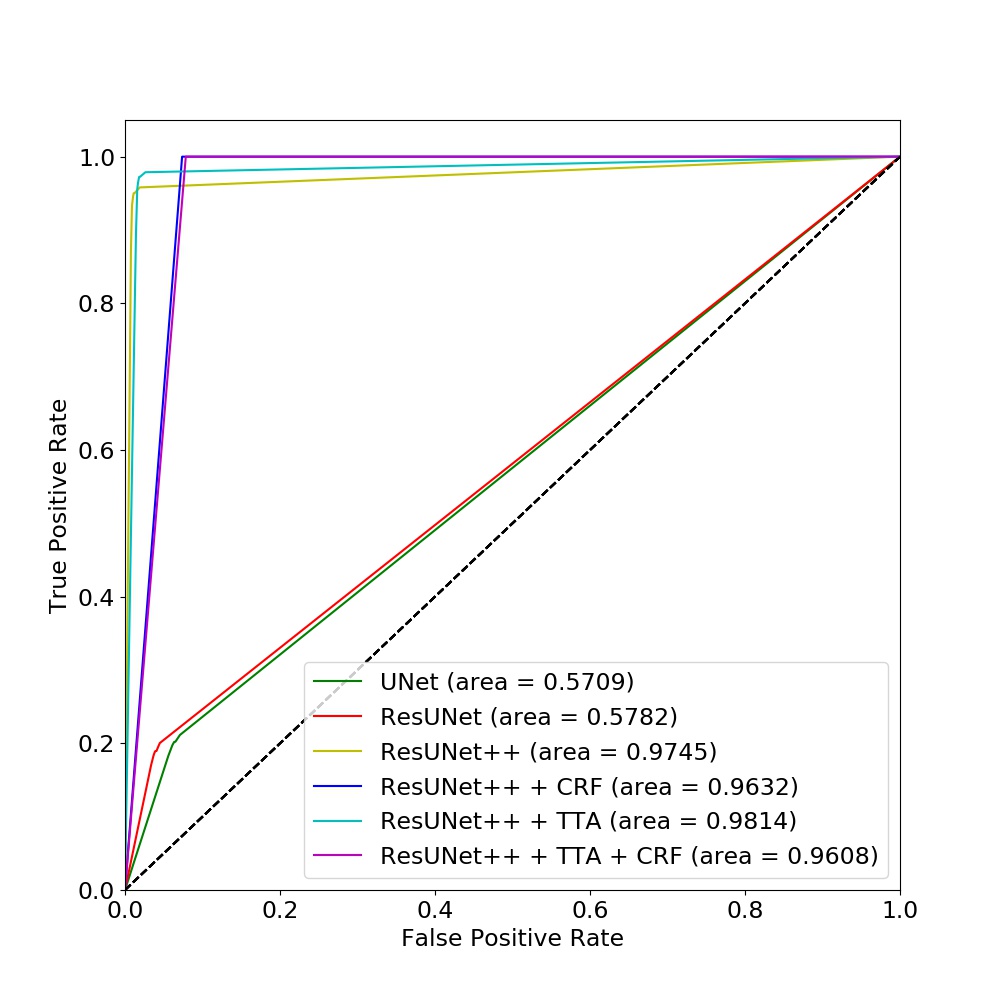}}
    \caption{\ac{ROC} curve for all the models trained and tested on CVC-ClinicDB} 
    \label{fig:figure7}
      \vspace{-3mm}
\end{figure}
\begin{table}[!t]
 \caption{Results comparison on CVC-ClinicDB}
    \label{table:resultcvc612}
    \scriptsize
    \centering
           \begin{tabular}{ l c c c c} 
                \toprule
                Method & DSC & mIoU & Recall & Precision\\ 
              \bottomrule
                MultiResUNet$^\diamond$~\cite{ibtehaz2020multiresunet}&  - & 0.8497 & - & -\\
                cGAN$^\dag$~\cite{poomeshwaran2019polyp} & 0.8848 & 0.8127 & - & -\\ 
                SegNet\cite{wang2018development} & - & - & 0.8824 & -\\ 
                FCN$^\bullet$~\cite{li2017colorectal} & - &- &0.7732 & \textbf{0.8999}\\ 
                CNN~\cite{nguyen2018colorectal} & (0.62-0.87) &- &- & -\\
                {MSPB$^\psi$~CNN}~\cite{banik2020multi} & 0.8130 &- &0.7860 & 0.8090\\
                UNet~\cite{ronneberger2015u} &  0.6419 & 0.4711 & 0.6756 & 0.6868 \\ 
               ResUNet~\cite{zhang2018road} & 0.4510 & 0.4570 & 0.5775 & 0.5614 \\
                {PraNet~\cite{fan2020pranet}} & 0.8980 &0.8400 & - & - \\
                
                ResUNet-mod~\cite{zhang2018road} & 0.7788 & 0.4545 & 0.6683 & 0.8877\\ 
               \resunetplusplus~\cite{jha2019resunet++} & 0.9199 & 0.8892 & 0.9391 & 0.8445 \\ 
                \resunetplusplus + \ac{CRF}  & \textbf{0.9203} & \textbf{0.8898} & \textbf{0.9393} & 0.8459\\ 
                \resunetplusplus + \ac{TTA} & 0.9020 & 0.8826 & 0.9065 & 0.8539 \\
                \resunetplusplus + \ac{TTA} + \ac{CRF} & 0.9017 & 0.8828 & 0.9060 & 0.8549\\
                 \bottomrule
                \multicolumn{4}{l}{$^\dag$ Conditional generative adversarial network} \hspace{.1cm} $^\diamond${Data augmentation}\\
	   \multicolumn{4}{l}{$^\bullet$Fully convolutional network \hspace{.1cm} {$^\psi$ multi-scale patch-based}}
\end{tabular}
 \vspace{-3mm} 
\end{table}

\subsection{Results comparison on CVC-ClinicDB}
CVC-ClinicDB is a commonly used dataset for polyp segmentation. Therefore, it becomes important that we bring different work from the literature together and compare the proposed algorithms with the existing works. We compare our algorithms with the~\ac{SOTA} algorithms. Table~\ref{table:resultcvc612} demonstrates that the combination of \resunetplusplus and \ac{CRF} achieves \ac{DSC} of 0.9293 and \ac{mIoU} of 0.8898, which is 2.23\% improvement on PraNet~\cite{fan2020pranet} in \ac{DSC} and 4.98\% improvement in \ac{mIoU}, respectively, and the proposed methods shows the \ac{SOTA} result on CVC-ClinicDB.

The \ac{ROC} curve measures the performance for the classification problem provided a set threshold. We have set the probability threshold of $0.5$. The combination of \resunetplusplus and \ac{TTA} has the maximum \ac{AUC-ROC} of 0.9814, as shown in Figure~\ref{fig:figure7}. Therefore, the results in Table~\ref{table:resultcvc612} and Figure~\ref{fig:figure7} show that applying \ac{TTA} gives an improvement on CVC-ClinicDB. 

\begin{table}[!t]
 \caption{Results comparison on CVC-ColonDB}
    \label{table:resultCVC-ColonDB}
\scriptsize
    \centering
           \begin{tabular}{ l c c c c c } 
                \toprule
                Method & DSC & \ac{mIoU} & Recall & Precision\\ 
              \bottomrule
                {FCN-8S + Otsu}~\cite{akbari2018polyp} & 0.8100 & - &0.7480 & -\\ 
                {FCN-8s + Texton}~\cite{zhang2017automated} & 0.7014 & - &0.7566 &-\\
                {SA-DOVA Descriptor}~\cite{bernal2012towards}& 0.5533& -& 0.6191& -\\
                {PraNet~\cite{fan2020pranet}} & 0.7090 & 0.6400 & - & - \\
                
                \resunetplusplus~\cite{jha2019resunet++}& 0.8469 &0.8456 &\textbf{0.8511} &0.8003  \\
                {\resunetplusplus + CRF}& 0.8458 &0.8456 &0.8497 &0.7767\\ %
                {\resunetplusplus + TTA}& \textbf{0.8474} &\textbf{0.8466} &0.8434 &0.8118\\ %
                {\resunetplusplus + TTA + CRF}&0.8452 &0.8459 &0.8411 &\textbf{0.8125}\\ 
                \bottomrule
                  \vspace{-5mm} 
\end{tabular}
\end{table}	

\subsection{Results comparison on CVC-ColonDB dataset}
Our results using the CVC-ColonDB dataset are presented in Table \ref{table:resultCVC-ColonDB}. The table shows that proposed method of combining ResUNet++ and TTA achieved the highest \ac{DSC} of 0.8474, which is 3.74\% higher than \ac{SOTA}~\cite{akbari2018polyp}, and \ac{mIoU} of 0.8466 which is 20.66\% higher than~\cite{fan2020pranet}. The recall and precision of all three proposed methods are quite acceptable. When compared with ResUNet++, there is an improvement of 1.22\% in precision. There are negligible differences in recall, with ResUNet++ slightly outperforming the others.


\begin{table}[!t]
 \caption{Results on ETIS-Larib Polyp DB} 
    \label{table:resultetislarib}
    \scriptsize
    \centering
           \begin{tabular}{l c c c c c} 
                \toprule
                Method & DSC & \ac{mIoU} & Recall & Precision\\ 
              \bottomrule
              {PraNet~\cite{fan2020pranet}} & 0.6280 & 0.5670 & - & - \\
                ResUNet++~\cite{jha2019resunet++} &\textbf{0.6364} &\textbf{0.7534} &\textbf{0.6346} &0.6467\\
                {ResUNet++ + CRF} &0.6228 &0.7520 &0.6242 &0.5648\\
                {ResUNet++ + TTA}&0.6136  &0.7458 &0.5996 &\textbf{0.6565}\\ 
                {ResUNet++ + TTA + CRF}&0.6018 &0.7426 &0.5914 &0.5755\\ 
                \bottomrule
                 \vspace{-5mm} 
\end{tabular}
\end{table}						

\subsection{Results comparison on ETIS-Larib Polyp DB}
Table~\ref{table:resultetislarib} shows the results of the proposed models on the ETIS-Larib Polyp DB. In this case, we do not compare the results with UNet and ResUNet, but compare the models directly with \resunetplusplus as it already showed superior performance on Kvasir-SEG and CVC-ClinicDB~\cite{jha2019resunet++}. Here, there are only marginal differences in the results of \resunetplusplus, ``\resunetplusplus + CRF", ``\resunetplusplus + TTA", and ``\resunetplusplus + CRF + TTA". However, \resunetplusplus achieves maximum \ac{DSC} of 0.6364, which is 0.84\% improvement over \ac{SOTA}~\cite{fan2020pranet} and \ac{mIoU} of 0.7534 which is 18.64\% improvement over~\cite{fan2020pranet}. The recall of \resunetplusplus is 0.6346, which is slightly higher than the proposed methods. However, the precision of combining \resunetplusplus and \ac{TTA} is higher as compared to \resunetplusplus.

From the results, we can say that the performance of architecture is data specific. Our proposed methods outperformed \ac{SOTA} over five independent datasets, however, \resunetplusplus shows better results than the combinational approaches on ETIS-Larib dataset. Still, the precision of combining \resunetplusplus and \ac{TTA} is slightly higher than \resunetplusplus. It is to be noted that ETIS-Larib contains only $196$ images, out of which only $156$ images are used for training. Even with the small training dataset, the models are performing satisfactory as compared to the \ac{SOTA}~\cite{fan2020pranet} with significant margin in \ac{mIoU}, which can be considered as the strength of the algorithm. 


\begin{table}[!t]
\caption{Results on Kvasir-Sessile} 
    \label{table:resultkvasirsessile}
    \scriptsize
    \centering
           \begin{tabular}{l c c c c c} 
                \toprule
                Method & DSC & \ac{mIoU} & Recall & Precision\\ 
              \bottomrule
                ResUNet++~\cite{jha2019resunet++} &0.4600 &0.64086  &0.4382  &0.5838 \\
                {ResUNet++ + CRF} &0.4522 &0.6394 &0.4326 &0.5708 \\
                {ResUNet++ + TTA}&\textbf{0.5042} &\textbf{0.6606} &\textbf{0.4851} &\textbf{0.6796} \\ 
                {ResUNet++ + TTA + CRF}&0.4901 &0.6565&0.4766  &0.6277\\ 
                \bottomrule
                 \vspace{-6mm} 
\end{tabular}
\end{table}

\vspace{-1mm}
\subsection{Results on Kvasir-Sessile}
\vspace{-1mm}
As this is the first work on Kvasir-Sessile, we have compared the proposed methods with \resunetplusplus. Table~\ref{table:resultkvasirsessile} shows that combining \resunetplusplus and \ac{TTA} gives the \ac{DSC} of 0.5042, \ac{mIoU} of 0.6606 which can be considered a decent score on a smaller size dataset. The dataset contains small, diverse images, which are difficult to generalize with very few training samples. 

\vspace{-2mm}

\begin{table}[t!]
 \caption{Results comparison on CVC-VideoClinicDB}
    \label{table:CVC-VideoClinicDB}
    \scriptsize
    \centering
           \begin{tabular}{ l c c c c c } 
                \toprule
                Method & DSC & \ac{mIoU} & Recall & Precision\\ 
              \bottomrule
                \resunetplusplus~\cite{jha2019resunet++} &0.8798 &0.8730 &\textbf{0.7749} &0.6702 \\
                {\resunetplusplus + CRF} &\textbf{0.8811} &\textbf{0.8739} &0.7743 &\textbf{0.6706}\\ %
                {\resunetplusplus + TTA} &0.8125 &0.8467 &0.6896 &0.6421\\ %
                {\resunetplusplus + TTA + CRF} &0.8130 &0.8477 &0.6875 &0.6276\\ 
                \bottomrule
\end{tabular}
 \vspace{-6mm} 
\end{table}						

\vspace{1mm}
\subsection{Results comparison on CVC-VideoClinicDB}
Table~\ref{table:CVC-VideoClinicDB} shows the results of the proposed models on the CVC-VideoClinicDB. From the results, we can see that all models perform well on the dataset despite the fact that masks are not pixel perfect. One of the reasons for high performance is the presence of  $11,954$ polyps and normal video frames that was used in training and testing. The combination of ResUNet++ and CRF obtained a DSC of $0.8811$, \ac{mIoU} of $0.8739$, recall of $0.7743$, and precision of $0.6706$ which is quite acceptable for the segmentation task with this type of dataset. In CVC-VideoClinicDB, the ground-truth is marked with a oval or circle shape. However, it is understandable that pixel-precise annotations of this dataset will need great manual effort from expert endoscopists and engineers.  
\vspace{-2mm}

\begin{table}[t!]
 \caption{Results comparison on ASUMayo Clinic}
    \label{table:ASUMayo}
\scriptsize
    \centering

           \begin{tabular}{ l c c c c c } 
                \toprule
                Method & DSC & \ac{mIoU} & Recall & Precision\\ 
              \bottomrule
                \resunetplusplus~\cite{jha2019resunet++} &0.8743 &0.8569 &\textbf{0.6534}  &0.4896    \\
                {\resunetplusplus + CRF} &\textbf{0.8850}  &\textbf{0.8635} &0.6504 &0.4858    \\ %
                {\resunetplusplus + TTA} &0.8553 &0.8535    &0.6162 &\textbf{0.4912} \\ %
                {\resunetplusplus + TTA + CRF} &0.8550 &0.8551 &0.6107 &0.4743    \\ 
                \bottomrule
\end{tabular}
 \vspace{-4mm} 
\end{table}

\subsection{Results comparison on AUS-Mayo ClinicDB}
Table~\ref{table:ASUMayo} shows the results of the proposed models on the ASU-Mayo ClinicDB. ASU-Mayo contains 18,781 frames, both polyp and non-polyp images. The combination of ResUNet++ and CRF obtained a DSC of 0.8850 and \ac{mIoU} of 0.8635. As in the real clinical settings, the models trained on this type of dataset are more meaningful (as it contains both polyp and non-polyp frames). The capability to achieve good performance for these more challenging datasets is one of the strengths of the proposed method. This is supported by the fact that this dataset also contains a sufficient amount of images to enable sufficient training.



\begin{table}[t!]
 \caption{Results comparison using (Kvasir-SEG + CVC-ClinicDB) as the training set}
    \label{table:mixeddataset}
    \scriptsize
    \centering
          \begin{tabular}{@{} c l c c c c @{}}  
                \bottomrule
                Test set & Method & DSC & mIoU &Recall &Precision\\ 
                \toprule
                \multirow{4}{*}{\rotatebox[origin=c]{90}{\parbox{1cm}{CVC\-ColonDB}}}
                &\resunetplusplus~\cite{jha2019resunet++} &0.4974 &0.6800 & 0.4787  &\textbf{0.6019}    \\
                &{\resunetplusplus + CRF} & 0.4920 &0.6788 &0.4744 &0.5636   \\ %
                &{\resunetplusplus + TTA} & \textbf{0.5084} &\textbf{0.6859} &\textbf{0.4795} &0.5973   \\ %
                &{\resunetplusplus + TTA + CRF}&0.5061  &0.6852 &0.4775 &0.5770\\ 
                \bottomrule
                \multirow{4}{*}{\rotatebox[origin=c]{90}{\parbox{1cm}{CVC-Video\-ClinicDB}}}
                
                &\resunetplusplus~\cite{jha2019resunet++} &0.3460 & 0.6348 & \textbf{0.2272}  & \textbf{0.3383}\\
                &{\resunetplusplus + CRF} & 0.3552 & 0.6412 & 0.2228 & 0.3065\\ %
                &{\resunetplusplus + TTA} & 0.3573 & 0.6440 & 0.2104 & 0.3338 \\ %
                &{\resunetplusplus + TTA + CRF}& \textbf{0.3603} & \textbf{0.6468} & 0.2068 & 0.3038 \\ 
                \bottomrule 
\end{tabular}
 \vspace{-5mm} 
\end{table}

\subsection{Results comparison on mixed dataset}
To check the performance of the proposed approaches on the images captured using different devices, we have mixed the Kvasir-SEG and CVC-ClinicDB and used them for training. The model were tested on CVC-ColonDB and CVC-VideoClinicDB. Table~\ref{table:mixeddataset} shows the result of the mixed dataset on both datasets. The combination of ResUNet++ and TTA obtains a \ac{DSC} of 0.5084 and \ac{mIoU} of 0.6859 with CVC-ColonDB. The combination of ResUNet++, CRF, and TTA obtained a \ac{DSC} of 0.3603 and \ac{mIoU} of 0.6468 with CVC-VideoClinicDB. 

From the table, we can see that the combination of ResUNet++, CRF, and TTA performs better or very competitive in both still images and video frames. Here, it is also evident that the model trained on the smaller dataset (Kvasir-SEG and CVC-ClinicDB) which do not include non-polyp images is not performing well on larger and diverse datasets (CVC-VideoClinicDB) that contain both polyp and non-polyp frames.  Additionally, for the CVC-VideoClinicDB datasets, the provided ground truth is not perfect (oval/circle) shaped. As the model trained on Kvasir-SEG and CVC-ClinicDB have perfect annotations, the model is good at predicting a perfect shaped mask. When we make predictions on the CVC-VideoClinicDB with imperfect masks, even if the predictions are good, the scores may not be high because of the difference in the provided ground truth and the predicted masks. 

\begin{table}[t]
 \caption{Cross-dataset results using \textbf{Kvasir-SEG} as the training set}
    \label{table:crossdatakvasir}
   \scriptsize
    \centering
           \begin{tabular}{@{} c l c c c c @{}} 
                \bottomrule
                Test set & Method & DSC & \ac{mIoU} & Recall & Precision\\ 
                \toprule
            \multirow{4}{*}{\rotatebox[origin=c]{90}{\parbox{1cm}{CVC-ClinicDB}}}    
            
              &\resunetplusplus~\cite{jha2019resunet++} & 0.6468 & 0.7311 & 0.6984 & 0.6510\\ 
              &{\resunetplusplus + CRF} & 0.6458 & 0.7321 & 0.6955 & 0.6425\\ 
              &{\resunetplusplus + TTA} & \textbf{0.6737} & \textbf{0.7507 }& \textbf{0.7108} & \textbf{0.6833}\\ 
              &{\resunetplusplus + TTA + CRF} & 0.6712 & 0.7506 & 0.7078 & 0.6680\\ 

              \toprule 
            \multirow{4}{*}{\rotatebox[origin=c]{90}{\parbox{1cm}{ETIS-Larib Polyp DB}}}    
                &\resunetplusplus~\cite{jha2019resunet++}& \textbf{0.4017} &0.6415 &\textbf{0.4412} &0.3925\\ 
                &{\resunetplusplus+ CRF}& 0.4012 &0.6427  &0.4379 &0.3755\\
                &{\resunetplusplus + TTA}& 0.4014 &\textbf{0.6468} &0.4294 &\textbf{0.4014}\\
                &{\resunetplusplus + TTA + CRF}&0.3997 &0.6466 &0.4267 &0.3710\\ 

                 \toprule
            \multirow{4}{*}{\rotatebox[origin=c]{90}{\parbox{1cm}{CVC-ColonDB}}}    
                &\resunetplusplus~\cite{jha2019resunet++} &0.5135 &0.6742 & 0.5398 &0.5461\\ %
                &{\resunetplusplus + CRF} &0.5122 & 0.6748 &0.5367 &0.5285\\ %
                &{\resunetplusplus + TTA}& \textbf{0.5593} &\textbf{0.7030 }&\textbf{0.5626} &\textbf{0.5944}\\ %
                &{\resunetplusplus + TTA + CRF}& 0.5563 &0.7024 &0.5595 &0.5811\\ 
               
                \toprule
            \multirow{4}{*}{\rotatebox[origin=c]{90}{\parbox{1cm}{CVC-Video\-ClinicDB}}}    
               &\resunetplusplus~\cite{jha2019resunet++}&0.3175 &0.6082 &\textbf{0.2915} &0.3299 \\ 
              &{\resunetplusplus + CRF}&0.3334 &0.6185 &0.2862 &0.3141\\ 
              &{\resunetplusplus + TTA}&0.3505 &0.6337 &0.2601&\textbf{0.3488}\\ 
              &{\resunetplusplus  + TTA + CRF}&\textbf{0.3601} &\textbf{0.6402 }&0.2555 &0.3252\\ 
              
               \bottomrule
            \multirow{4}{*}{\rotatebox[origin=c]{90}{\parbox{1cm}{ASU-Mayo}}}    
              &{{\resunetplusplus~\cite{jha2019resunet++}}}&0.3482&0.6346 &\textbf{0.2196} &\textbf{0.2021}\\ 
              &{\resunetplusplus + CRF}&0.3747  &0.6516 &0.2136 &0.1797 \\ 
              &{\resunetplusplus + TTA}&0.3823 &0.6583 &0.1962 &0.2165\\ 
              &{\resunetplusplus  + TTA + CRF}&\textbf{0.3950} &\textbf{0.6681} &0.1890 &0.1781\\ 

 \bottomrule
\end{tabular}
  \vspace{-4mm}
\end{table}	

\vspace{-1mm}
\subsection{Cross-dataset result evaluation on Kvasir-SEG}
For the cross-dataset evaluation, we trained the models on the Kvasir-SEG dataset and tested it on the other five independent datasets. Table~\ref{table:crossdatakvasir} shows the results of cross-data generalizability of \resunetplusplus alone, and with the \ac{CRF} and \ac{TTA} techniques. The results of the models trained on Kvasir-SEG produces an average best \ac{mIoU} of 0.6817 and an average best \ac{DSC} of 0.4779 for both image and video datasets. From the above table, we can see that the proposed combinational approaches are performing competitive. For the image datasets, the combination of ResUNet++ and TTA is performing better, and for the video datasets, the combination of ResUNet++, CRF, and TTA is performing best. It is to be noted that we are training a model with 1000 Kvasir-SEG pixel segmented polyps and testing on (for example, 11,954 frames) oval-shaped polyp ground truth. Here, even if the predictions are correct, the evaluation scores will not be good because of the oval/circle shaped ground truth. Moreover, the datasets such as ASU-Mayo and CVC-VideoClinicDB are heavily imbalanced, but the model trained on Kvasir-SEG contains at least one polyp. This may also have caused the poor performance. 




\begin{table}[t]
 \caption{Cross-dataset results on \textbf{CVC-ClinicDB} as the training set}
    \label{table:crossdataCVCclinicDB}
  \scriptsize
    \centering
          \begin{tabular}{@{} c l c c c c @{}}  
                \bottomrule
                Test set & Method & DSC & mIoU &Recall &Precision\\ 
                \toprule
                
                \multirow{4}{*}{\rotatebox[origin=c]{90}{\parbox{1cm}{Kvasir-SEG}}}
              &\resunetplusplus~\cite{jha2019resunet++}&0.6876 &0.7374 &0.7027 &0.7354\\ 
              &{\resunetplusplus + CRF}&0.6877 &0.7389 &0.7004 &0.7371 \\ 
              &{\resunetplusplus + TTA}&\textbf{0.7218} &0.7616 &\textbf{0.7225} &\textbf{0.7855}\\ 
              &{\resunetplusplus + TTA + CRF}&0.7208 &\textbf{0.7621} &0.7204 &0.7831\\
                  \toprule
             \multirow{4}{*}{\rotatebox[origin=c]{90}{\parbox{1cm}{CVC-ColonDB}}} 
           
               & \resunetplusplus~\cite{jha2019resunet++} &0.5489  &0.6942 &0.5577 &0.5816\\ 
                &{\resunetplusplus + CRF}& 0.5470 &0.6949 &0.5546 &0.5727 \\ 
               & {\resunetplusplus + TTA}&\textbf{0.5686} &0.7080  &\textbf{0.5702} &\textbf{0.5935}\\ 
                &{\resunetplusplus + + TTA + CRF}& 0.5667 &\textbf{0.7081} &0.5687 &0.5773\\
                \bottomrule
                  \multirow{6}{*}{\rotatebox[origin=c]{90}{\parbox{1cm}{ETIS-Larib Polyp DB}}} 

             & FCN-VGG~\cite{brandao2017fully} &0.7023 &0.5420  &- &- \\ 
              & {\resunetplusplus~\cite{jha2019resunet++}} &0.4012 &0.6398 &\textbf{0.4232}&0.4013\\ 
              & {\resunetplusplus + CRF} &0.3990 &0.6403 &0.4191 &0.3974\\ 
                &{\resunetplusplus + TTA} &0.4027 &\textbf{0.6522 } &0.3969 &\textbf{0.4235}\\ 
                &{\resunetplusplus + TTA + CRF} &0.3973 &0.6514 &0.3906 &0.4078 \\
                \bottomrule
                
                \multirow{4}{*}{\rotatebox[origin=c]{90}{\parbox{1cm}{CVC-Video\-ClinicDB}}} 
              &{\resunetplusplus~\cite{jha2019resunet++}} &0.3666 &0.6422 &\textbf{0.2568} &0.3632\\
             & {\resunetplusplus + CRF}&0.3788 & 0.6500 & 0.2530 & 0.3399 \\ 
             & {\resunetplusplus + TTA}&0.3941 &0.6582 &0.2516 &\textbf{0.3829}\\ 
             & {\resunetplusplus + TTA + CRF}&\textbf{0.3988} &\textbf{0.6616} &0.2481 &0.3542\\ 
                \bottomrule
              \multirow{4}{*}{\rotatebox[origin=c]{90}{\parbox{1cm}{ASU-Mayo}}}
             & {{\resunetplusplus~\cite{jha2019resunet++}}} &0.2797 &0.6113 &\textbf{0.1627 }&0.1443 \\  
             & {\resunetplusplus + CRF} &0.3167 &0.6323 &0.1591  &0.1348\\ 
             & {\resunetplusplus + TTA} &0.3085&0.6331 &0.1265 &\textbf{0.1571}\\ 
             & {\resunetplusplus + TTA + CRF} &\textbf{0.3233} &\textbf{0.6426} &0.1225 &0.1270\\
 \bottomrule
\end{tabular}
  \vspace{-4mm}
\end{table}		


\subsection{Cross-dataset evaluation on CVC-ClinicDB}
\vspace{-1mm}
To further test generaliziblity, we trained the models on CVC-CliniDB and tested it across five independent, diverse image and video datasets. Tables~\ref{table:crossdataCVCclinicDB} shows the results of cross-data generalizability. Like the previous test on Kvasir-SEG, the results follow the same pattern with the combination of ResUNet++ and TTA outperforming others on the image datasets and the combination of ResUNet++, CRF, and TTA outperforming its competitors on video datasets. ResUNet++ and TTA still remain competitive. Moreover, the values of \ac{DSC} and \ac{mIoU} of the best model are similar for both the CVC-VideoClinicDB and the ASU-Mayo Clinic dataset. We have compared the results with the existing work that used  CVC-CliniDB for training and ETIS-Larib for testing. Our model achieves highest \ac{mIoU} of 0.6522. 


\subsection{Result summary}
In summary, from all obtained results (i.e., qualitative, quantitative, and \ac{ROC} curve), the following main observations can be drawn: 
(i) the proposed \resunetplusplus is capable of segmenting the smaller, larger and regular polyps; (ii) the combination of \resunetplusplus with \ac{CRF} achieves the best performance in terms of \ac{DSC}, \ac{mIoU}, recall and precision when trained and tested on the same dataset (see Table \ref{table:resultcvc612}, Table \ref{table:CVC-VideoClinicDB}, and Table \ref{table:ASUMayo}) whereas it remains competitive when tested on other datasets; (iii) the combination of \resunetplusplus and TTA and the combination of \resunetplusplus, CRF and TTA performs similar for the mixed datasets; (iv) the combination of \resunetplusplus and TTA outperforms others on still images; (v) the combination of \resunetplusplus, CRF and TTA shows improvement on all the video datasets compared to \resunetplusplus; (vi) all the models perform better when the images have higher contrast; (vii) \resunetplusplus is particularly good at segmenting smaller and flat or sessile polyps, which is a prerequisite for developing an ideal \ac{CADx} polyp detection system~\cite{jha2019resunet++}; (viii) \resunetplusplus fails especially on the images that contains over-exposed regions termed as saturation or contrast (see Figure~\ref{fig:samekvasirfail}); (ix) ResUNet and ResUNet-mod particularly showed over-segmented or under-segmented results, (see Figure~\ref{fig:flatpolypkvasir}).  


\section{Discussion}
\label{sec:discussion}
\begin{table}[t!]
\caption{Total number of trainable parameters}
\label{table:trainable}
\scriptsize
\centering
\begin{tabular}{l c}
\toprule
 Model & Trainable parameters \\ 
 \bottomrule
U-Net & 5,400,289\\ 
ResUNet & 8,221,121\\ 
ResUNet-mod & 2,058,465\\ 
\resunetplusplus & 16,228,001\\ 
\bottomrule
\end{tabular}
\end{table}
\subsection{General Performance}
The tables and figures suggest that applying \ac{CRF} and \ac{TTA} improved the performance of \resunetplusplus on the same datasets, mixed datasets and cross-datasets. Specifically, the combination of \resunetplusplus and \ac{TTA}, and the combination of  \resunetplusplus, CRF and \ac{TTA} are more generalizable for all the datasets, where TTA with ResUNet++ performs best on the still images, and the combinations of ResUNet++, CRF, and TTA are outperforming others on video datasets. 
For all of the proposed models, the value of AUC is greater than $0.93$. This indicates that our models are good at distinguishing between the polyp and non-polyps. It also suggests that the model produces sufficient sensitivity.

The total number of trainable parameters increases by increasing the number of blocks in the networks (see Table~\ref{table:trainable}). However, in \resunetplusplus, there is significant performance gain that compensates for the training time, and our model requires fewer parameters if we compare with the models that use pre-trained encoders. 


\subsection{Cross Dataset Performance}
The cross-data test is an excellent technique to determine the generalizing capability of a model. The presented work is an initiative towards improving the generalizability of segmentation methods. Our contribution towards generalizability is to train on one dataset and test on several other public datasets that may come from different centers and use different scope manufacturers. Thus, we believe that to tackle this issue, out-of-sample multicenter data must be used to test the built methods. The work is a step forward in raising an issue regarding method interpretability and we also raise questions about generalizability and domain adaptation of supervised methods in general.

From the results analyses, we can see that different proposed algorithms perform well with different types of datasets. For instance, \ac{CRF} outperformed others on tables~\ref{table:resultcvc612}, \ref{table:CVC-VideoClinicDB}, and \ref{table:ASUMayo}. \ac{TTA} showed improvement on tables~\ref{table:resultCVC-ColonDB},~\ref{table:mixeddataset}, \ref{table:crossdatakvasir} and \ref{table:crossdataCVCclinicDB}. \ac{CRF} performs better than TTA while trained and tested on video datasets (see tables~\ref{table:CVC-VideoClinicDB} and \ref{table:ASUMayo}). \ac{CRF} also outperformed \ac{TTA} on most of the images dataset. However, \ac{TTA} still remains competitive. On the mixed dataset and the cross-dataset test, \ac{TTA} performs better than \ac{CRF} on all the datasets. On the mixed datasets and on the cross-dataset test on videos, the combination of \resunetplusplus, CRF, and TTA remains the best choice (see tables \ref{table:mixeddataset}, \ref{table:crossdatakvasir}, and \ref{table:crossdataCVCclinicDB}). There is a performance improvement over \resunetplusplus while combining CRF,  TTA, and the combination of CRF and  TTA.

However, there is no significant performance improvement of any methods on the others. From the results, we can see that the results are typically data-dependent. However, as the proposed methods perform well on video frames, it may work better in the clinic, as the output from a colonoscope is a video stream. Thus, it becomes critical to show the results with all three approaches on each dataset. Therefore, we provide extensive experiments showing both success (Figure~\ref{fig:flatpolypkvasir}, Figure~\ref{fig:cross-data612kvasir}) and failure cases (Figure~\ref{fig:samekvasirfail}) and present the overall analysis.


\subsection{Challenges}
There are several challenges associated with segmenting polyps, such as bowel-quality preparation during colonoscopy, angle of the cameras, superfluous information, and varying morphology, which can affect the overall performance of a \ac{DL} model. For some of the images, there even exists variation in the decision between endoscopists. While \resunetplusplus with \ac {CRF} and \ac{TTA} also struggle with producing satisfactory segmentation maps for these images, it performs considerably better than our previous model and also outperforms another \ac{SOTA} algorithm. 

The quality of a colonoscopy examination is largely determined by the experience and skill of the endoscopist~\cite{yamada2019development}. Our proposed model can help in two ways: (i) it can be used to segment a detected polyp, providing an extra pair of eyes to the endoscopist; and (ii) it performs well on both flat and small polyps, which are often missed during endoscopic examinations. The qualitative analysis (see Figure~\ref{fig:flatpolypkvasir}) and the quantitative analyses from the above tables and figures support this argument. This is a major strength of our work and makes it a candidate for clinical testing.

\subsection{Possible Limitations}
Possible limitations of this work are that it is a retrospective study. Prospective clinical evaluation is essential because data analyzed with the retrospective study is the different prospective study (for example, the case of missing data that should be considered on the basis of best-case and worse case scenarios)~\cite{mori2018detecting}. Also, all data in these experiments are curated, while a prospective clinical trial would mean testing on full colonscopy videos. During model training, we have resized all the images to $256\times256$ to reduce the complexity, which costs in loss of information, and can affect the overall performance. We have worked on optimizing the code, but further optimization may exist, that can potentially improve the performance of the model.

\vspace{5mm}
\section{Conclusion}
\label{sec:conclusion}

In this paper, we have presented the \resunetplusplus architecture for semantic polyp segmentation. We took inspiration from the residual block, \ac{ASPP}, and attention block to design the novel \resunetplusplus architecture. Furthermore, we applied \ac{CRF} and \ac{TTA} to improve the results even more. We have trained and validated the combination of \resunetplusplus with \ac{CRF} and \ac{TTA} using six publicly available datasets, and analyzed and compared the results with the \ac{SOTA} algorithm on specific datasets. Moreover, we analyzed the cross-data generalizability of the proposed model towards developing generalizable semantic segmentation models for automatic polyp segmentation. A comprehensive evaluation of the proposed model trained and tested on six different datasets showed good performance of the (\resunetplusplus and \ac{CRF}) on image datasets and  (\resunetplusplus and \ac{TTA}), (\resunetplusplus, \ac{CRF}, and \ac{TTA})  model for the mixed datasets and cross-datasets. Further, a detailed study on cross-dataset generalizability of the models trained on Kvasir-SEG and CVC-ClinicDB and tested on five independent datasets, confirmed the robustness of the proposed  \resunetplusplus + \ac{TTA} method for cross-dataset evaluation.

The strength of our method is that we successfully detected smaller and flat polyps, which are usually missed during colonoscopy examination~\cite{leufkens2012factors,wang2018development}. Our model can also detect the polyps that would be difficult for the endoscopists to identify without careful investigations. Therefore, we believe that the \resunetplusplus architecture, along with the additional  \ac{CRF} and \ac{TTA} steps, could be one of the potential areas to investigate, especially for the overlooked polyps. We also point out that the lack of generalization issues of the models, which is evidenced by the unsatisfactory result for cross-dataset evaluation in most of the cases. In the future, our \ac{CADx} system should also be investigated on other bowel conditions. Moreover, a prospective trial should also be conducted with image and video datasets. 


\section*{Acknowledgement}
This work is funded in part by Research Council of Norway project number 263248. Experiments are performed on the Experimental Infrastructure for Exploration of Exascale Computing (eX3), supported by the Research Council of Norway under contract 270053.
\bibliographystyle{IEEEtran}
\bibliography{references.bib} 

\begin{thebibliography}{10}
\providecommand{\url}[1]{#1}
\csname url@samestyle\endcsname
\providecommand{\newblock}{\relax}
\providecommand{\bibinfo}[2]{#2}
\providecommand{\BIBentrySTDinterwordspacing}{\spaceskip=0pt\relax}
\providecommand{\BIBentryALTinterwordstretchfactor}{4}
\providecommand{\BIBentryALTinterwordspacing}{\spaceskip=\fontdimen2\font plus
\BIBentryALTinterwordstretchfactor\fontdimen3\font minus
  \fontdimen4\font\relax}
\providecommand{\BIBforeignlanguage}[2]{{%
\expandafter\ifx\csname l@#1\endcsname\relax
\typeout{** WARNING: IEEEtran.bst: No hyphenation pattern has been}%
\typeout{** loaded for the language `#1'. Using the pattern for}%
\typeout{** the default language instead.}%
\else
\language=\csname l@#1\endcsname
\fi
#2}}
\providecommand{\BIBdecl}{\relax}
\BIBdecl

\bibitem{jha2019resunet++}
D.~Jha \emph{et~al.}, ``Resunet++: An advanced architecture for medical image
  segmentation,'' in \emph{Proc. of IEEE ISM.}, 2019, pp. 225--230.

\bibitem{bray2018global}
F.~Bray, J.~Ferlay, I.~Soerjomataram, R.~L. Siegel, L.~A. Torre, and A.~Jemal,
  ``Global cancer statistics 2018: Globocan estimates of incidence and
  mortality worldwide for 36 cancers in 185 countries,'' \emph{CA: a cancer
  journal for clinicians}, vol.~68, no.~6, pp. 394--424, 2018.

\bibitem{matsuda2017advances}
T.~Matsuda, A.~Ono, M.~Sekiguchi, T.~Fujii, and Y.~Saito, ``Advances in image
  enhancement in colonoscopy for detection of adenomas,'' \emph{Nat. Revi.
  Gastroenter. \& Hepato.}, vol.~14, no.~5, pp. 305--314, 2017.

\bibitem{jha2020kvasir}
D.~Jha \emph{et~al.}, ``Kvasir-seg: A segmented polyp dataset,'' in \emph{Proc.
  of MMM}, 2020, pp. 451--462.

\bibitem{ahn2012miss}
S.~B. Ahn, D.~S. Han, J.~H. Bae, T.~J. Byun, J.~P. Kim, and C.~S. Eun, ``The
  miss rate for colorectal adenoma determined by quality-adjusted, back-to-back
  colonoscopies,'' \emph{Gut and liver}, vol.~6, no.~1, pp. 64--70, 2012.

\bibitem{heresbach2008miss}
D.~o. Heresbach, ``Miss rate for colorectal neoplastic polyps: a prospective
  multicenter study of back-to-back video colonoscopies,'' \emph{Endoscopy},
  vol.~40, no.~04, pp. 284--290, 2008.

\bibitem{zimmermann2019right}
Zimmermann-Fraedrich \emph{et~al.}, ``Right-sided location not associated with
  missed colorectal adenomas in an individual-level reanalysis of tandem
  colonoscopy studies,'' \emph{Gastroenterology}, vol. 157, no.~3, pp.
  660--671, 2019.

\bibitem{shaukat2015longer}
A.~Shaukat \emph{et~al.}, ``Longer withdrawal time is associated with a reduced
  incidence of interval cancer after screening colonoscopy,''
  \emph{Gastroenterology}, vol. 149, no.~4, pp. 952--957, 2015.

\bibitem{vazquez2017benchmark}
D.~V{\'a}zquez \emph{et~al.}, ``A benchmark for endoluminal scene segmentation
  of colonoscopy images,'' \emph{Journal of healthcare engineering}, vol. 2017,
  2017.

\bibitem{tobiasinstrument2020}
T.~Roß \emph{et~al.}, ``Robust medical instrument segmentation challenge
  2019,'' \emph{arXiv preprint arXiv:2003.10299v1}, 2020.

\bibitem{bernal2015wm}
J.~Bernal, F.~J. S{\'a}nchez, G.~Fern{\'a}ndez-Esparrach, D.~Gil,
  C.~Rodr{\'\i}guez, and F.~Vilari{\~n}o, ``Wm-dova maps for accurate polyp
  highlighting in colonoscopy: Validation vs. saliency maps from physicians,''
  \emph{Computeri. Med. Imag. and Graph.}, vol.~43, pp. 99--111, 2015.

\bibitem{bernal2012towards}
J.~Bernal, J.~S{\'a}nchez, and F.~Vilarino, ``Towards automatic polyp detection
  with a polyp appearance model,'' \emph{Patt. Recognit.}, vol.~45, no.~9, pp.
  3166--3182, 2012.

\bibitem{silva2014toward}
J.~Silva, A.~Histace, O.~Romain, X.~Dray, and B.~Granado, ``Toward embedded
  detection of polyps in wce images for early diagnosis of colorectal cancer,''
  \emph{Int. Jour. of Comput. Assis. Radiol. and Surg.}, vol.~9, no.~2, pp.
  283--293, 2014.

\bibitem{tajbakhsh2015automated}
N.~Tajbakhsh, S.~R. Gurudu, and J.~Liang, ``Automated polyp detection in
  colonoscopy videos using shape and context information,'' \emph{IEEE Trans.
  Med. Imag.}, vol.~35, no.~2, pp. 630--644, 2015.

\bibitem{angermann2017towards}
Q.~Angermann \emph{et~al.}, ``Towards real-time polyp detection in colonoscopy
  videos: Adapting still frame-based methodologies for video sequences
  analysis,'' in \emph{Comput. Assis. and Robot. Endos. and Clin. Image-Based
  Proced.}, 2017, pp. 29--41.

\bibitem{bernal2018polyp}
J.~Bernal \emph{et~al.}, ``Polyp detection benchmark in colonoscopy videos
  using gtcreator: A novel fully configurable tool for easy and fast annotation
  of image databases,'' in \emph{Proceedings of CARS conference}, 2018.

\bibitem{pozdeev2019automatic}
A.~A. Pozdeev, N.~A. Obukhova, and A.~A. Motyko, ``Automatic analysis of
  endoscopic images for polyps detection and segmentation,'' in \emph{Proc. of
  EIConRus}, 2019, pp. 1216--1220.

\bibitem{bernal2017comparative}
J.~Bernal \emph{et~al.}, ``Comparative validation of polyp detection methods in
  video colonoscopy: results from the miccai 2015 endoscopic vision
  challenge,'' \emph{IEEE Trans. Med. Imag.}, vol.~36, no.~6, pp. 1231--1249,
  2017.

\bibitem{akbari2018polyp}
M.~Akbari \emph{et~al.}, ``Polyp segmentation in colonoscopy images using fully
  convolutional network,'' in \emph{Proc. of EMBC}, 2018, pp. 69--72.

\bibitem{wang2018development}
P.~Wang \emph{et~al.}, ``Development and validation of a deep-learning
  algorithm for the detection of polyps during colonoscopy,'' \emph{Nat.
  biomed. engineer.}, vol.~2, no.~10, pp. 741--748, 2018.

\bibitem{badrinarayanan2017segnet}
V.~Badrinarayanan, A.~Kendall, and R.~Cipolla, ``Segnet: A deep convolutional
  encoder-decoder architecture for image segmentation,'' \emph{IEEE trans. on
  patt. analys. and mach. intellige.}, vol.~39, no.~12, pp. 2481--2495, 2017.

\bibitem{guo2019giana}
Y.~B. Guo and B.~Matuszewski, ``Giana polyp segmentation with fully
  convolutional dilation neural networks,'' in \emph{Proc. of VISIGRAPP}, 2019,
  pp. 632--641.

\bibitem{yamada2019development}
M.~Yamada \emph{et~al.}, ``Development of a real-time endoscopic image
  diagnosis support system using deep learning technology in colonoscopy,''
  \emph{Scienti. repo.}, vol.~9, no.~1, pp. 1--9, 2019.

\bibitem{poomeshwaran2019polyp}
J.~Poomeshwaran, K.~S. Santhosh, K.~Ram, J.~Joseph, and M.~Sivaprakasam,
  ``Polyp segmentation using generative adversarial network,'' in \emph{Proc.
  of EMBC}, 2019, pp. 7201--7204.

\bibitem{kang2019ensemble}
J.~Kang and J.~Gwak, ``Ensemble of instance segmentation models for polyp
  segmentation in colonoscopy images,'' \emph{IEEE Access}, vol.~7, pp.
  26\,440--26\,447, 2019.

\bibitem{ali2019endoscopy}
S.~Ali \emph{et~al.}, ``Endoscopy artifact detection (ead 2019) challenge
  dataset,'' \emph{arXiv preprint arXiv:1905.03209}, 2019.

\bibitem{nguyen2019robust}
N.-Q. Nguyen and S.-W. Lee, ``Robust boundary segmentation in medical images
  using a consecutive deep encoder-decoder network,'' \emph{IEEE Access},
  vol.~7, pp. 33\,795--33\,808, 2019.

\bibitem{de2019training}
V.~de~Almeida~Thomaz, C.~A. Sierra-Franco, and A.~B. Raposo, ``Training data
  enhancements for robust polyp segmentation in colonoscopy images,'' in
  \emph{Proc. of CBMS}, 2019, pp. 192--197.

\bibitem{sun2019colorectal}
X.~Sun, P.~Zhang, D.~Wang, Y.~Cao, and B.~Liu, ``Colorectal polyp segmentation
  by u-net with dilation convolution,'' \emph{arXiv preprint arXiv:1912.11947},
  2019.

\bibitem{jha2020doubleu}
D.~Jha, M.~A. Riegler, D.~Johansen, P.~Halvorsen, and H.~D. Johansen,
  ``Doubleu-net: A deep convolutional neural network for medical image
  segmentation,'' in \emph{Proc. of IEEE CBMS}, 2020.

\bibitem{ibtehaz2020multiresunet}
N.~Ibtehaz and M.~S. Rahman, ``Multiresunet: Rethinking the u-net architecture
  for multimodal biomedical image segmentation,'' \emph{Neural Networks}, vol.
  121, pp. 74--87, 2020.

\bibitem{brandao2018towards}
P.~Brandao \emph{et~al.}, ``Towards a computed-aided diagnosis system in
  colonoscopy: automatic polyp segmentation using convolution neural
  networks,'' \emph{Jour. of Medi. Robot. Resear.}, vol.~3, no.~02, p. 1840002,
  2018.

\bibitem{deng2009imagenet}
J.~Deng, W.~Dong, R.~Socher, L.-J. Li, K.~Li, and L.~Fei-Fei, ``Imagenet: A
  large-scale hierarchical image database,'' in \emph{Proc. of CVPR}, 2009, pp.
  248--255.

\bibitem{zhang2018road}
Z.~Zhang, Q.~Liu, and Y.~Wang, ``Road extraction by deep residual u-net,''
  \emph{IEEE Geosci. and Remo. Sens. Lett.}, vol.~15, no.~5, pp. 749--753,
  2018.

\bibitem{ronneberger2015u}
O.~Ronneberger, P.~Fischer, and T.~Brox, ``U-net: Convolutional networks for
  biomedical image segmentation,'' in \emph{Proc. of MICCAI}, 2015, pp.
  234--241.

\bibitem{hu2018squeeze}
J.~Hu, L.~Shen, and G.~Sun, ``Squeeze-and-excitation networks,'' in \emph{Proc.
  of CVPR}, 2018, pp. 7132--7141.

\bibitem{chen2017rethinking}
L.-C. Chen, G.~Papandreou, F.~Schroff, and H.~Adam, ``Rethinking atrous
  convolution for semantic image segmentation,'' \emph{arXiv preprint
  arXiv:1706.05587}, 2017.

\bibitem{vaswani2017attention}
A.~Vaswani, N.~Shazeer, N.~Parmar, J.~Uszkoreit, L.~Jones, A.~N. Gomez,
  {\L}.~Kaiser, and I.~Polosukhin, ``Attention is all you need,'' in
  \emph{Proc. of NIPS}, 2017, pp. 5998--6008.

\bibitem{ioffe2015batch}
S.~Ioffe and C.~Szegedy, ``Batch normalization: Accelerating deep network
  training by reducing internal covariate shift,'' \emph{arXiv preprint
  arXiv:1502.03167}, 2015.

\bibitem{lecun2015deep}
Y.~LeCun, Y.~Bengio, and G.~Hinton, ``Deep learning,'' \emph{nature}, vol. 521,
  no. 7553, pp. 436--444, 2015.

\bibitem{he2016deep}
K.~He, X.~Zhang, S.~Ren, and J.~Sun, ``Deep residual learning for image
  recognition,'' in \emph{Proc. of CVPR}, 2016, pp. 770--778.

\bibitem{wang2019nested}
L.~Wang, R.~Chen, S.~Wang, N.~Zeng, X.~Huang, and C.~Liu, ``Nested dilation
  network (ndn) for multi-task medical image segmentation,'' \emph{IEEE
  Access}, vol.~7, pp. 44\,676--44\,685, 2019.

\bibitem{he2016identity}
K.~He, X.~Zhang, S.~Ren, and J.~Sun, ``Identity mappings in deep residual
  networks,'' in \emph{Proc. of ECCV}, 2016, pp. 630--645.

\bibitem{chen2014semantic}
L.-C. Chen, G.~Papandreou, I.~Kokkinos, K.~Murphy, and A.~L. Yuille, ``Semantic
  image segmentation with deep convolutional nets and fully connected crfs,''
  \emph{arXiv preprint arXiv:1412.7062}, 2014.

\bibitem{chen2017deeplab}
L.~Chen, G.~Papandreou, I.~Kokkinos, K.~Murphy, and A.~L. Yuille, ``Deeplab:
  Semantic image segmentation with deep convolutional nets, atrous convolution,
  and fully connected crfs,'' \emph{IEEE trans. on pattern anal. and mach.
  intelli.}, vol.~40, no.~4, pp. 834--848, 2017.

\bibitem{chen2016attention}
L.-C. Chen, Y.~Yang, J.~Wang, W.~Xu, and A.~L. Yuille, ``Attention to scale:
  Scale-aware semantic image segmentation,'' in \emph{Proc. of CVPR}, 2016, pp.
  3640--3649.

\bibitem{wang2018deep}
Y.~Wang \emph{et~al.}, ``Deep attentional features for prostate segmentation in
  ultrasound,'' in \emph{Proc. of MICCAI}, 2018, pp. 523--530.

\bibitem{nie2018asdnet}
D.~Nie, Y.~Gao, L.~Wang, and D.~Shen, ``Asdnet: Attention based semi-supervised
  deep networks for medical image segmentation,'' in \emph{Proc. of MICCAI},
  2018, pp. 370--378.

\bibitem{sinha2019multi}
A.~Sinha and J.~Dolz, ``Multi-scale guided attention for medical image
  segmentation,'' \emph{arXiv preprint arXiv:1906.02849}, 2019.

\bibitem{alam2018conditional}
F.~I. Alam, J.~Zhou, A.~W.-C. Liew, X.~Jia, J.~Chanussot, and Y.~Gao,
  ``Conditional random field and deep feature learning for hyperspectral image
  classification,'' \emph{IEEE Trans. on Geosci. and Remo. Sens.}, vol.~57,
  no.~3, pp. 1612--1628, 2018.

\bibitem{pogorelov2017kvasir}
K.~Pogorelov \emph{et~al.}, ``Kvasir: A multi-class image dataset for computer
  aided gastrointestinal disease detection,'' in \emph{Proc. of MMSYS}, 2017,
  pp. 164--169.

\bibitem{chollet2015keras}
F.~Chollet \emph{et~al.}, ``Keras,'' 2015.

\bibitem{abadi2016tensorflow}
M.~Abadi \emph{et~al.}, ``Tensorflow: A system for large-scale machine
  learning,'' in \emph{Proc. of OSDI}, 2016, pp. 265--283.

\bibitem{li2017colorectal}
Q.~Li \emph{et~al.}, ``Colorectal polyp segmentation using a fully
  convolutional neural network,'' in \emph{Proc. of CISP-BMEI}, 2017, pp. 1--5.

\bibitem{nguyen2018colorectal}
Q.~Nguyen and S.-W. Lee, ``Colorectal segmentation using multiple
  encoder-decoder network in colonoscopy images,'' in \emph{Proc. of IKE},
  2018, pp. 208--211.

\bibitem{banik2020multi}
D.~Banik, D.~Bhattacharjee, and M.~Nasipuri, ``A multi-scale patch-based deep
  learning system for polyp segmentation,'' in \emph{Advan. Comput. and Syst.
  for Secur.}, 2020, pp. 109--119.

\bibitem{fan2020pranet}
D.-P. Fan \emph{et~al.}, ``Pranet: Parallel reverse attention network for polyp
  segmentation,'' in \emph{Proc. of MICCAI}, 2020, pp. 263--273.

\bibitem{zhang2017automated}
L.~Zhang, S.~Dolwani, and X.~Ye, ``Automated polyp segmentation in colonoscopy
  frames using fully convolutional neural network and textons,'' in \emph{Proc.
  ov MIUA}, 2017, pp. 707--717.

\bibitem{brandao2017fully}
P.~Brandao \emph{et~al.}, ``Fully convolutional neural networks for polyp
  segmentation in colonoscopy,'' in \emph{Medical Imaging 2017: Computer-Aided
  Diagnosis}, vol. 10134, 2017, pp. 101\,340F1 -- 101\,340F1.

\bibitem{mori2018detecting}
Y.~Mori and S.-e. Kudo, ``Detecting colorectal polyps via machine learning,''
  \emph{Nat. biomed. engineer.}, vol.~2, no.~10, pp. 713--714, 2018.

\bibitem{leufkens2012factors}
A.~Leufkens, M.~Van~Oijen, F.~Vleggaar, and P.~Siersema, ``Factors influencing
  the miss rate of polyps in a back-to-back colonoscopy study,''
  \emph{Endoscopy}, vol.~44, no.~05, pp. 470--475, 2012.

\end{thebibliography}
\end{document}